\documentclass{article}

    \PassOptionsToPackage{numbers, compress}{natbib}
\usepackage[main, final]{neurips_2026}

\usepackage[utf8]{inputenc} %
\usepackage[T1]{fontenc}    %
\usepackage{hyperref}       %
\usepackage{url}            %
\usepackage{booktabs}       %
\usepackage{amsmath}
\usepackage{amssymb}
\usepackage{amsfonts}       %
\usepackage{nicefrac}       %
\usepackage{microtype}      %
\usepackage{xcolor}         %
\usepackage{graphicx} 
\usepackage{tikz}
\usepackage{enumitem}
\usepackage{listings}       %
\lstdefinestyle{prompt}{basicstyle=\ttfamily\scriptsize, columns=fullflexible, keepspaces=true, breaklines=true, breakindent=0pt, frame=single, framesep=3pt, xleftmargin=3pt, xrightmargin=3pt, aboveskip=6pt, belowskip=6pt, upquote=true, showstringspaces=false}

\newcommand{\gact}{d_{\mathrm{act}}}

\newcommand{\Tspan}{\mathcal{T}}

\definecolor{takeawaybg}{RGB}{249,240,228}
\definecolor{takeawayborder}{RGB}{143,93,48}
\newcounter{takeaway}
\newcommand{\takeaway}[1]{%
  \refstepcounter{takeaway}%
  \par\vspace{2pt}%
  \noindent\begin{tikzpicture}
    \node[
      draw=takeawayborder,
      fill=takeawaybg,
      rounded corners=2pt,
      line width=0.4pt,
      inner xsep=4pt,
      inner ysep=2.5pt,
      text width=\dimexpr\linewidth-9pt\relax,
      align=left
    ] {\small\textbf{Takeaway~\thetakeaway.} #1};
  \end{tikzpicture}%
  \par\vspace{1pt}%
}
\hypersetup{
    colorlinks=true,
    linkcolor=[rgb]{0.1,0.1,0.5},   
    citecolor=[rgb]{0.1,0.1,0.5},
    urlcolor=[rgb]{0.1,0.1,0.5}
}

\title{\textsc{Say the Same, Act Differently:}\\ Diagnosing and Steering Text--Action Divergence in Reasoning VLAs}

\author{%
  \textbf{Zihao Feng}$^{1}$ \quad
  \textbf{Qingzhao Zhang}$^{2}$ \quad
  \textbf{Chunyu Xia}$^{1}$ \\
  \textbf{Bo Yu}$^{3}$ \quad
  \textbf{Z. Morley Mao}$^{1}$ \quad
  \textbf{Ramesh Govindan}$^{1}$ \\
  {\normalfont $^{1}$University of Southern California \quad $^{2}$University of Arizona \quad $^{3}$General Motors} \\
  {\normalfont \texttt{\{zihaofen, chunyuxi, morley.mao, ramesh\}@usc.edu}} \\
  {\normalfont \texttt{qzzhang@arizona.edu} \quad \texttt{bo.3.yu@gm.com}}
}

\begin{document}

\maketitle

\begin{abstract}
Reasoning vision-language-action (VLA) models pair physical control with natural-language rationales that supervisors can read to monitor embodied safety.
Prior attacks on VLA reasoning corrupt the rationale itself and leave visible evidence in the text.
We show that visual perturbations can instead alter the physical action while keeping the generated rationale word-for-word identical.
At the shared text/action interface of Alpamayo for driving and InstructVLA for manipulation, our text-neutral action-subspace (TNAS) diagnostic shows that the locally most action-sensitive direction is nearly orthogonal to the active text-loss gradients, with $1.87\%$ and $5.28\%$ norm overlap.
At equal hidden-state perturbation norms across both models, interventions along this text-neutral residual shift actions $4.64\times$ to $47.8\times$ more than text-gradient perturbations, whereas text-gradient perturbations induce $21.2\times$ to $277\times$ more text drift.
Steering visual perturbations along this direction reaches the same decoupling from the image.
Across $154$ obstacle-annotated driving scenes in Alpamayo, TNAS-guided attacks induce a new collision under a word-for-word identical rationale in $52.6\%$ of scenes under full-image perturbations and $35.1\%$ under a localized forward-camera patch, compared with $4.55\%$ for unguided action attacks.
Relative to a text-preserving baseline, TNAS guidance raises exact rationale preservation from $83.0\%$ to $89.5\%$ and cuts teacher-forced text drift by $68.4\%$ at a comparable collision rate.
In InstructVLA, TNAS guidance cuts text drift by $40.5\%$ and preserves the exact response on $85.0\%$ of examples alongside a $1.73$-unit action-chunk shift.
An unchanged rationale therefore cannot certify a safe action, so monitors for reasoning VLAs must verify the predicted action channel directly.
\end{abstract}

\section{Introduction}

Vision-language-action (VLA) models translate visual observations and language instructions into embodied actions.
Reasoning VLAs additionally produce natural-language scene descriptions, rationales, or plans alongside continuous action predictions~\citep{tian2024drivevlm,jiang2024senna,zawalski2024robotic,wang2025alpamayo,yang2025instructvla}.
Unlike latent representations or raw motor commands, this language channel gives human supervisors an explicit, readable account of what the model perceives and intends to do.
A supervisor can inspect that account for a missed obstacle, an incorrect target object, or an intention that conflicts with the instruction, potentially before the action is executed.
The same idea drives chain-of-thought monitoring for language agents~\citep{baker2025monitoring,korbak2025monitorability}, and for embodied models the stakes are even higher because an erroneous physical motion can cause collisions or physical damage.
Whether this interface can be trusted for safety hinges on one question: \emph{can the action become unsafe without any change in what the model says?}

\begin{figure}[t]
    \centering
    \includegraphics[width=0.95\linewidth]{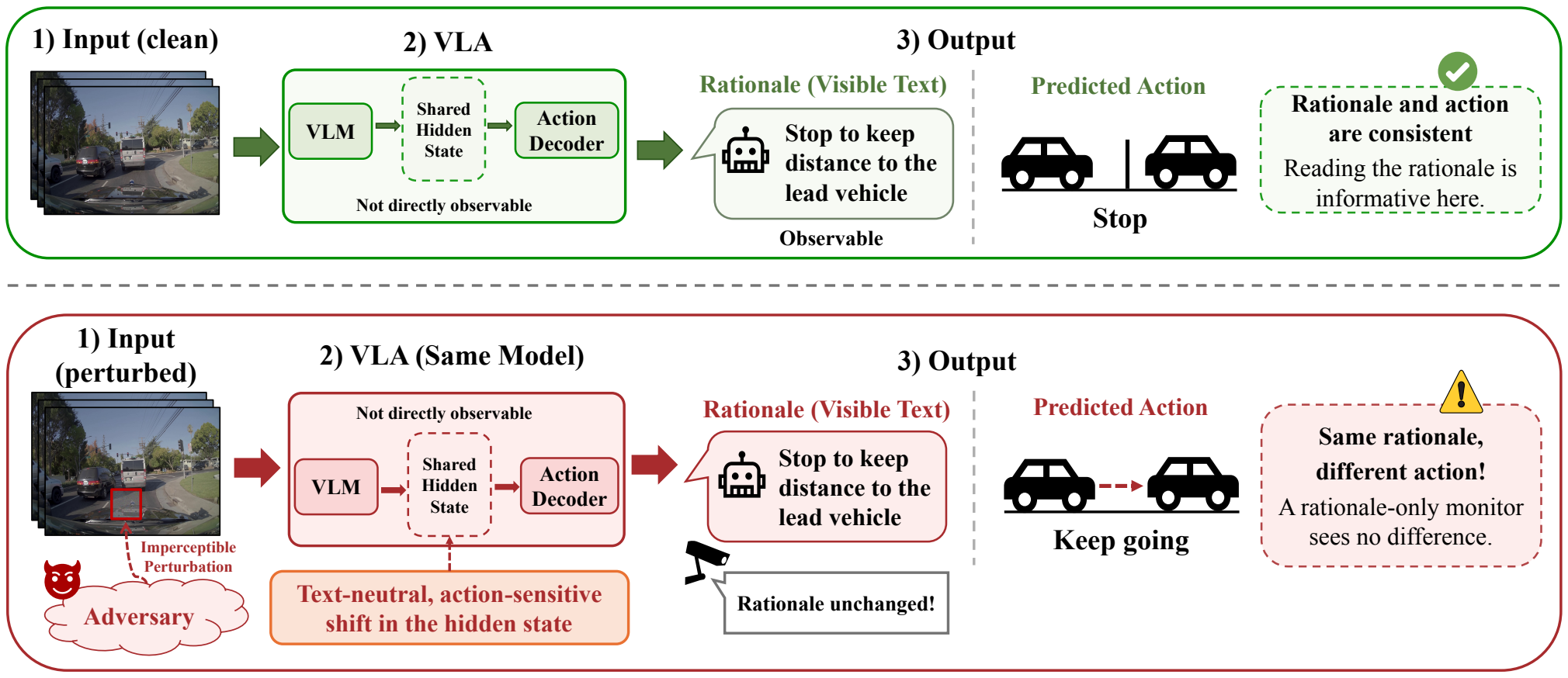}
    \caption{\textbf{Rationale-preserving action divergence in reasoning VLAs.} On a clean driving input, the VLM and action decoder read the same unobservable hidden state to produce a matching rationale and stopping action. Adding an imperceptible perturbation to the input shifts that shared hidden state along a text-neutral, action-sensitive direction. The model still outputs ``Stop to keep distance to the lead vehicle'' word for word, but its predicted action flips from stopping to continuing into the lead vehicle.}
    \label{fig:monitorability_gap}
\end{figure}

Recent attacks on reasoning VLAs redirect behavior by changing the reasoning itself.
TRAP steers chain-of-thought and downstream actions toward an adversary's target with adversarial patches~\citep{huang2026trap}, while \citet{trinh2026altered} corrupt the intermediate reasoning trace directly, for example by substituting object names.
Both leave evidence in the channel a monitor reads. Because the reasoning text has changed, an incorrect object reference or an instruction-inconsistent plan is visible to a supervisor.
In contrast, as illustrated in \hyperref[fig:monitorability_gap]{Figure~\ref*{fig:monitorability_gap}}, we study \emph{rationale-preserving action divergence}, where a perturbed input changes the predicted action while leaving the generated rationale word-for-word identical to the clean run.
When a perturbed driving input produces a colliding plan under the exact same rationale as the clean input, a monitor restricted to that rationale receives identical evidence in both cases.

This failure mode is rooted in how reasoning and action share a representation.
In the reasoning VLAs studied here, action prediction reads continuous hidden states that also produce the text~\citep{wang2025alpamayo,yang2025instructvla}.
Because the rationale is a discrete, many-to-one readout of those continuous states, many distinct hidden states can map to the same generated text.
Under ordinary inputs the text and action co-vary, which makes rationales informative in typical operation.
For a rationale monitor to certify safety, however, any unsafe change in the action must also change the rationale.
Prior reasoning-hijacking attacks show that altering the text alters the action, leaving open the converse question: does the shared hidden state contain action-sensitive variation that leaves the text readout unchanged, and can visual inputs reach it?

We answer this question with the \textsc{text-neutral action-subspace} (TNAS) diagnostic and a corresponding input-level attack.
At the shared text--action interface, TNAS identifies the locally most action-sensitive direction in the hidden state, measures its overlap with the active text-loss gradients, and projects out that text-gradient component to isolate a text-neutral action direction ($d_{\mathrm{TNAS}}$).
We first perturb the clean hidden state directly along $d_{\mathrm{TNAS}}$ and control directions, under both fixed text-drift budgets and equal perturbation norms, to verify that this local direction shifts the predicted action while leaving the token distribution nearly unchanged.
We then align image perturbations with $d_{\mathrm{TNAS}}$ during input-level optimization to steer the model toward silent action divergence under paired free generation.
Across both Alpamayo for autonomous driving and InstructVLA for robotic manipulation, we find the same geometric decoupling and summarize our main contributions as follows:

\noindent\textbf{Contributions.}
\begin{itemize}[itemsep=1.5pt, parsep=0pt, topsep=1pt, partopsep=0pt, leftmargin=*]
    \item \textbf{Rationale-preserving action divergence through visual inputs.} 
    We demonstrate that visual perturbations can substantially alter a reasoning VLA's action while preserving the exact generated rationale. 
    In Alpamayo, TNAS-guided perturbations introduce a geometric collision absent from the clean plan while leaving the rationale word-for-word unchanged in $52.6\%$ of annotated scenes under full-image perturbations and $35.1\%$ under a localized forward-camera patch, and in InstructVLA, they preserve the exact response in $85.0\%$ of examples with a mean action-chunk shift of $1.73$ normalized units.
    \item \textbf{A local geometric mechanism for text--action decoupling.} 
    We introduce TNAS and show that the action-sensitive direction at the shared interface overlaps the active text-loss span by only $1.87\%$ of its norm in Alpamayo and $5.28\%$ in InstructVLA. 
    Under equal perturbation norms across both models, perturbing the hidden state along the projected text-neutral direction shifts actions $4.64\times$ to $47.8\times$ more than text-gradient perturbations, whereas text-gradient perturbations induce $21.2\times$ to $277\times$ more teacher-forced KL.
    \item \textbf{Mechanism-guided attacks with stronger text preservation at the same level of action deviation.} 
    Directly combining an action-deviation loss and a text-preservation penalty creates competing gradients that interfere with each other during optimization. 
    Adding TNAS alignment to the attack objective resolves this conflict by steering hidden-state updates along the text-neutral direction: at the strongest guidance weights, exact rationale preservation rises from $83.0\%$ to $89.5\%$ in Alpamayo and from $77.5\%$ to $85.0\%$ in InstructVLA, while teacher-forced text drift falls by $68.4\%$ and $40.5\%$ at an unchanged driving collision rate.
\end{itemize}

\section{\textsc{Text-Neutral Action-Subspace} Diagnostic}
\label{sec:mechanism}

\subsection{Setup and probe construction}
\label{sec:metrics}
\label{sec:setup}
\label{sec:method}

\paragraph{Models and cohorts.}
We evaluate Alpamayo\label{sec:alpamayo_internal}, an autonomous driving VLA, on $200$ clips and InstructVLA\label{sec:instructvla}, a robotic manipulation VLA, on $200$ Bridge-distribution examples, both fixed as described in \hyperref[app:dataset]{Appendix~\ref*{app:dataset}}.
Alpamayo is our primary safety case study because its predicted trajectory can be scored against logged traffic actors on $154$ obstacle-annotated clips.
With InstructVLA, we test whether the same geometry holds in a second embodiment that uses a flow-matching action expert and concise task-intention responses.
\paragraph{Shared-state hook and intervention coordinates.}
\label{sec:method_shared_state}
For a block of $K$ transformer layers of hidden width $d$ and an active reasoning- or response-token span $R$, an additive intervention $\delta\in\mathbb{R}^{Kd}$ broadcasts one vector per layer over $R$. 
A usable block must expose both a nonzero text basis and a strong action interface in the same state. 
As detailed in \hyperref[app:layer]{Appendix~\ref*{app:layer}}, selecting the qualifying block with the strongest action response gives reasoning-token layers 32--35 in Alpamayo and response-token layers 14--17 in InstructVLA.
The pooled state $h\in\mathbb{R}^{Kd}$ concatenates per-layer token means on $R$ and supplies the input-level alignment readout. 

\paragraph{Action and text readouts.}
As detailed in \hyperref[app:action_metrics]{Appendix~\ref*{app:action_metrics}}, under the hooked forward pass, both action heads decode with a fixed 10-step Euler flow-matching solver and paired initial noise: Alpamayo integrates acceleration--curvature controls into 2D planar waypoints $\tau_t(\delta)\in\mathbb{R}^2$, and InstructVLA predicts normalized 7-DoF end-effector actions $a_t(\delta)\in\mathbb{R}^7$ (3D position, 3D rotation, and gripper). 
We measure action change as the mean displacement from the clean output ($\delta=0$), over $T=64$ waypoints at $10$ Hz ($6.4$ s, in meters) for Alpamayo and over $H=16$ chunk steps for InstructVLA:
\begin{equation}
    \Delta_{\mathrm{ADE}}(\delta)
    =
    \frac{1}{T}\sum_{t=1}^{T}
    \left\lVert
    \tau_t(\delta)-\tau_t(0)
    \right\rVert_2,
    \qquad
    \Delta_{\mathrm{IVLA}}(\delta)
    =
    \frac{1}{H}\sum_{t=1}^{H}
    \left\lVert
    a_t(\delta)-a_t(0)
    \right\rVert_2.
    \label{eq:action_shift}
\end{equation}
These \emph{fixed-prefix readouts} hold the clean rationale or response fixed as the text prefix, so the paired clean and perturbed passes differ only through the intervention.
As defined in \hyperref[app:tf_kl]{Appendix~\ref*{app:tf_kl}}, text drift is the forward teacher-forced KL divergence, from the clean to the perturbed next-token distribution, averaged over $R$.

\paragraph{Action-Jacobian direction.}
\label{sec:method_action_dir}
The action-Jacobian direction $\gact$ is a hidden-state direction to which the action output is locally sensitive. 
We do not take the gradient of $\Delta_{\mathrm{ADE}}$ or $\Delta_{\mathrm{IVLA}}$ directly. Both are norms of a displacement that is zero at $\delta=0$, so neither has a well-defined gradient at the clean point.
We instead work with the Jacobian of the action output itself.
Let $A_{\mathrm{hook}}(\delta)$ be the flattened action output of the hooked pass, namely the $64$ XY waypoints for Alpamayo and the $16\times7$ action chunk for InstructVLA, with local Jacobian
\begin{equation}
    J_A=\left.\frac{\partial A_{\mathrm{hook}}(\delta)}{\partial\delta}\right|_{\delta=0}.
\end{equation}
As detailed in \hyperref[app:local_probe_impl]{Appendix~\ref*{app:local_probe_impl}}, we obtain $\gact$ by backpropagating through the action readout to the hooked state and normalizing the resulting row-space direction of $J_A$. 
The same direction then goes through text-span residualization and the hidden-state intervention tests.

\paragraph{Active text-loss span.}
\label{sec:method_textspan}
The text side needs losses whose gradients with respect to $\delta$ are nonzero at the clean point $\delta=0$. 
Teacher-forced KL and clipped hinge losses do not qualify, because the KL gradient vanishes when the clean and perturbed distributions coincide and hinge losses saturate with zero gradient whenever the clean token already clears the margin threshold. 
We therefore capture local text sensitivity through two complementary losses that remain active at $\delta=0$, namely target negative log-likelihood (NLL), which tracks the probability mass assigned to the clean token across the vocabulary, and signed target-margin drop, which tracks the logit gap to the strongest competing token that governs top-1 token flips. 
For token $t\in R$ with clean target $y_t$ and logits $z_t$, let $p_t=\mathrm{softmax}(z_t)$ and $m_t=z_{t,y_t}-\max_{v\neq y_t}z_{t,v}$. 
Writing superscripts $\mathrm{cln}$ and $\mathrm{adv}$ for the clean and perturbed passes, the two losses are
\begin{equation}
    \mathcal{L}_{\mathrm{NLL}}(\delta)
    = \frac{1}{|R|}\sum_{t\in R} -\log p^{\mathrm{adv}}_{t,y_t},
    \qquad
    \mathcal{L}_{\mathrm{margin}}(\delta)
    = \frac{1}{|R|}\sum_{t\in R}\bigl(m^{\mathrm{cln}}_t-m^{\mathrm{adv}}_t\bigr).
    \label{eq:loss_nll}
\end{equation}
Gram--Schmidt on their clean-state gradients $g_{\mathrm{NLL}}$ and $g_{\mathrm{margin}}$ gives an orthonormal basis of the text-loss span $\Tspan=\mathrm{span}\{g_{\mathrm{NLL}},g_{\mathrm{margin}}\}$, which has rank 2 on every evaluated example.

\paragraph{TNAS residual.}
With $P_{\Tspan}$ the orthogonal projector onto $\Tspan$, the text-neutral action-subspace (TNAS) direction is
\begin{equation}
    d_{\mathrm{TNAS}}
    =
    \mathrm{normalize}\bigl((I-P_{\Tspan})\gact\bigr).
\end{equation}
By construction, $d_{\mathrm{TNAS}}$ is orthogonal to both text-loss gradients at $\delta=0$, so a small step along it leaves the NLL and margin of the clean rationale unchanged to first order while still moving the action.

\subsection{Clean-state geometry}

\paragraph{Probe 1. How much of the action does the text carry?}
\phantomsection
\label{sec:geometry}
\label{sec:ivla_geometry}
\label{sec:scene_semantics}
\label{sec:benign_reader}
\mbox{}\\
\textbf{Rationale-only action prediction}. To evaluate the consistency between the rationale and the action under ordinary inputs, we use a frozen Qwen3-14B model~\citep{yang2025qwen3}, configured as described in \hyperref[app:benign_reader]{Appendix~\ref*{app:benign_reader}}, to predict from the clean rationale alone whether Alpamayo's planned trajectory comes to rest within the $6.4$ s horizon.
Across the $1{,}181$ clips, where $24.5\%$ ($289/1{,}181$) of planned trajectories actually come to rest, the reader's prediction matches the trajectory outcome on $84.3\%$ of clips. Clips predicted to stop come to rest $63.6\%$ of the time, compared with $5.76\%$ for clips predicted to continue.
Even so, the text does not pin down the action. The $1{,}181$ clips share only $258$ distinct rationales, $36.9\%$ carry a rationale that also appears with the opposite stopping outcome, and the most frequent rationale (``Keep distance to the lead vehicle since it is directly ahead in our lane'') alone covers $245$ clips with $10.1$ m to $208$ m of forward displacement.
\mbox{}\\
\textbf{Hidden-state gradient projection}. The hidden-state geometry locates this gap at the shared interface.
To quantify how much of the action gradient lies in the text span, we project the action direction $\gact$ onto $\Tspan$ and divide by the norm of $\gact$, giving the projection-norm ratio $\rho=\lVert P_{\Tspan}\gact\rVert/\lVert\gact\rVert$.
Across the $200$ Alpamayo clips, $\rho$ averages $1.87\%$ (versus $\rho=1$ for text-gradient controls), roughly twice the $0.978\%$ mean of isotropic random unit vectors reported in \hyperref[app:projection]{Appendix~\ref*{app:projection}}, and across the $200$ InstructVLA examples, $\rho$ averages $5.28\%$.
Because the text-span projection $P_{\Tspan}\gact$ and its orthogonal complement $(I-P_{\Tspan})\gact$ are perpendicular, removing $\Tspan$ leaves $\sqrt{1-\rho^2}>99.8\%$ of $\gact$'s norm intact on average in both models, meaning that almost the entire action-sensitive direction is orthogonal to the text span.

\takeaway{\textbf{The rationale carries action information, but the text registers only a small part of what moves the action.} Read alone, Alpamayo's rationale predicts whether the plan stops on $84.3\%$ of clips, yet $36.9\%$ of clips share their rationale with the opposite outcome. At the shared interface, the projection-norm ratio $\rho$ of $\gact$ onto the active text span averages $1.87\%$ in Alpamayo and $5.28\%$ in InstructVLA, so projecting out the text span retains on average more than $99.8\%$ of the action direction's norm in both models.}

\subsection{Hidden-state intervention results}
\label{sec:method_kl}

\begin{figure}[t]
\centering
\includegraphics[width=\linewidth]{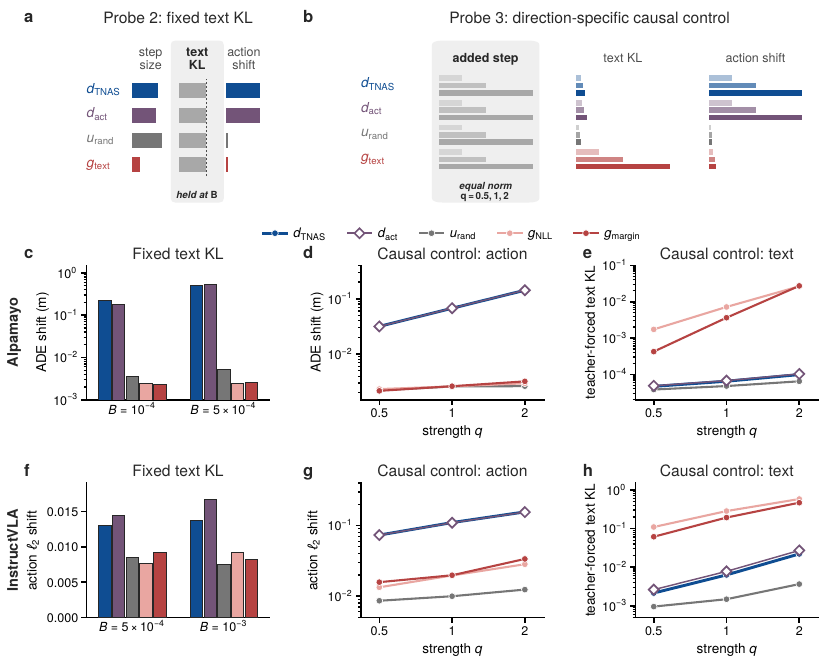}
\caption{\textbf{Hidden-state intervention probes.} \emph{Top:} schematic of the two protocols ($g_{\mathrm{text}}$ denotes $g_{\mathrm{NLL}}$ and $g_{\mathrm{margin}}$). (a)~Probe~2 adjusts each direction's step size so that all directions incur the same text-KL budget $B$ (shaded) and compares the resulting action shift. (b)~Probe~3 adds every direction at the same per-example norm $|q|S_{\mathrm{clip}}$ for $q\in\{0.5,1,2\}$ (shaded, light to dark) and compares the resulting text KL and action shift. \emph{Middle and bottom:} cohort means on $200$ Alpamayo clips and $200$ InstructVLA examples, with full values tabulated in \hyperref[app:probe_tables]{Appendix~\ref*{app:probe_tables}}. (c,f)~Action shift at fixed teacher-forced KL budgets $B$. (d,g)~Action shift and (e,h)~teacher-forced text KL across equal-norm strengths $q$, where $\gact$ (hollow diamonds) nearly overlaps $d_{\mathrm{TNAS}}$.}
\label{fig:probes}
\end{figure}

\hyperref[sec:geometry]{Probe~1} measures first-order gradients at $\delta=0$. 
To test whether this local separation holds when finite hidden-state perturbations pass through the nonlinear language and action heads, \hyperref[sec:matched_budget]{Probe~2} and \hyperref[sec:nulling_and_amplification]{Probe~3} compare $d_{\mathrm{TNAS}}$ and the unprojected action direction $\gact$ against three non-action control directions, namely the text-loss gradients ($g_{\mathrm{NLL}}$, $g_{\mathrm{margin}}$) and random unit vectors ($u_{\mathrm{rand}}$), on the same $200$ Alpamayo clips and $200$ InstructVLA examples. \hyperref[fig:probes]{Figure~\ref*{fig:probes}} summarizes both probes, and \hyperref[app:local_probe_impl]{Appendices~\ref*{app:local_probe_impl}} and~\ref{app:probe_tables} provide the implementation details and full tables. 
As shown in \hyperref[fig:probes]{Figure~\ref*{fig:probes}}, \hyperref[sec:matched_budget]{Probe~2} compares these directions under a fixed text-drift budget in panel~a, and \hyperref[sec:nulling_and_amplification]{Probe~3} compares them under a fixed hidden-state perturbation norm in panel~b.

\paragraph{Probe 2. Fixed text KL.}
\phantomsection
\label{sec:matched_budget}
\label{sec:alpamayo_matched_budget}
\label{sec:ivla_matched_budget}
To test how much action change a hidden-state perturbation can induce before noticeably altering the rationale, we hold teacher-forced text KL fixed across directions and compare the resulting action shifts. 
As illustrated in \hyperref[fig:probes]{Figure~\ref*{fig:probes}a,c,f}, for each direction, we select the step size on the perturbation grid whose teacher-forced KL matches a target budget $B$. 
\mbox{}\\
\textbf{Alpamayo}. Perturbing along $d_{\mathrm{TNAS}}$ shifts the planned trajectory by $0.227$ m at $B=10^{-4}$ and $0.504$ m at $B=5\times10^{-4}$, matching $\gact$ and exceeding the controls ($\le 5.10$ mm) by $63.3\times$ to $212\times$. 
\mbox{}\\
\textbf{InstructVLA}. Perturbing along $d_{\mathrm{TNAS}}$ and $\gact$ likewise produces the largest action shifts at both budgets, reaching $1.37\times10^{-2}$ and $1.66\times10^{-2}$ at $B=10^{-3}$ versus at most $9.18\times10^{-3}$ for the controls. 

\paragraph{Probe 3. Direction-specific causal control.}
\phantomsection
\label{sec:nulling_and_amplification}
\label{sec:amplification}
\label{sec:ivla_nulling_and_amplification}
\label{sec:ivla_amplification}
In \hyperref[sec:matched_budget]{Probe~2}, matching a fixed text-KL budget allows directions that barely perturb the text to take larger hidden-state steps than text-sensitive directions, as \hyperref[fig:probes]{Figure~\ref*{fig:probes}a} illustrates. 
To verify that $d_{\mathrm{TNAS}}$ is intrinsically more action-selective per unit norm rather than only benefiting from a larger step size, Probe~3 fixes the hidden-state perturbation magnitude across all directions, as depicted in \hyperref[fig:probes]{Figure~\ref*{fig:probes}b}. 
Starting from the clean state, Probe~3 adds each direction with the same per-example norm $|q|S_{\mathrm{clip}}$, where $S_{\mathrm{clip}}=|\langle h_{\mathrm{clean}},d_{\mathrm{TNAS}}\rangle|$ and $q\in\{0.5,1,2\}$, as detailed in \hyperref[app:sign_aligned]{Appendix~\ref*{app:sign_aligned}}. \hyperref[fig:probes]{Figure~\ref*{fig:probes}d,e,g,h} then shows how the action and text channels respond. 
Across both models, $d_{\mathrm{TNAS}}$ and the text-gradient controls show opposite selectivity, while $\gact$ lies on top of $d_{\mathrm{TNAS}}$ across all $q$. 
\mbox{}\\
\textbf{Alpamayo}. The trajectory shift along $d_{\mathrm{TNAS}}$ grows with $q$ to $0.142$ m at $q=2$, exceeding the text-gradient controls by $44.0\times$ to $47.8\times$ and the random control by $53.9\times$. 
Conversely, at $q=2$, $d_{\mathrm{TNAS}}$ induces only $9.87\times10^{-5}$ teacher-forced KL, whereas the text-gradient controls induce $269\times$ to $277\times$ higher KL. 
\mbox{}\\
\textbf{InstructVLA}. At $q=2$, $d_{\mathrm{TNAS}}$ shifts the action by $0.156$, exceeding the text-gradient controls by $4.64\times$ to $5.51\times$ and the random control by $12.6\times$, while keeping teacher-forced KL at $2.20\times10^{-2}$ compared with $21.2\times$ to $26.3\times$ higher KL for the text-gradient controls. 

\takeaway{\textbf{Hidden-state interventions along $d_{\mathrm{TNAS}}$ shift the action while barely moving the text.} Under a fixed text-KL budget, $d_{\mathrm{TNAS}}$ shifts Alpamayo's trajectory $63.3\times$ to $212\times$ more than the controls. Under an equal perturbation norm across both models, $d_{\mathrm{TNAS}}$ shifts the action $4.64\times$ to $47.8\times$ more than the text-gradient controls and $12.6\times$ to $53.9\times$ more than the random control, while the text-gradient controls induce $21.2\times$ to $277\times$ more teacher-forced KL.}

\section{TNAS-Guided Visual Perturbations}
\label{sec:attack}

\hyperref[sec:mechanism]{Section~\ref*{sec:mechanism}} showed that direct hidden-state interventions along $d_{\mathrm{TNAS}}$ shift the predicted action while leaving the next-token distribution nearly unchanged.
We now test whether an adversary modifying only the visual input can excite this same direction end-to-end in Alpamayo and InstructVLA, resolving the optimization conflict between altering the physical action and preserving the generated text.

\subsection{Attack setup}
\label{sec:attack_protocol}

\paragraph{Threat model.}
We consider white-box digital perturbations applied to the normalized, processed visual input tensor immediately before the visual encoder, with gradients backpropagated end-to-end through the vision encoder, language backbone, and action head.
In Alpamayo, full-image attacks perturb all $16$ camera-frame inputs across the four cameras and four temporal frames.
In InstructVLA, both full-image and localized attacks perturb the VLM/SigLIP visual stream.
\hyperref[sec:patch_attack]{Section~\ref*{sec:patch_attack}} further restricts the attacker to a spatially localized patch in each domain to mimic real-world scenarios where perturbations are attached to local objects or surfaces in the scene.

\paragraph{Optimization.}
Every attack runs $200$ steps of projected gradient descent (PGD)~\citep{goodfellow2015explaining,madry2018towards,zhang2022adversarial,zhang2024data} on its maximization objective $\mathcal{L}$,
\begin{equation}
    \delta^{(k+1)}
    =
    \Pi_{\mathcal{S}}\!\left[
        \delta^{(k)}
        +
        \alpha\,\mathrm{sign}\!\left(\nabla_{\delta}\mathcal{L}(\delta^{(k)})\right)
    \right],
    \label{eq:pgd_update}
\end{equation}
where $\Pi_{\mathcal{S}}$ projects onto the $\ell_\infty$ ball of radius $\varepsilon$ and the allowed spatial mask.
Full-image attacks use $\varepsilon=0.05$ and step size $\alpha=5\times10^{-4}$ in normalized pixel coordinates, whereas localized patches use a larger local budget ($\varepsilon=0.5$, $\alpha=5\times10^{-3}$) to compensate for their restricted spatial support.
Further implementation details on differentiable action rollouts are provided in \hyperref[app:input_protocols]{Appendix~\ref*{app:input_protocols}}.

\paragraph{Evaluation metrics.}
Our evaluation covers both controlled fixed-prefix readouts and end-to-end free generation, along with open-loop safety checks in Alpamayo:
\begin{itemize}[nosep, leftmargin=1.2em, label=\textbullet]
    \item \textbf{Fixed-prefix readouts.} Under the fixed clean text prefix, we report the \hyperref[sec:setup]{Section~\ref*{sec:setup}} readouts: action shift ($\Delta_{\mathrm{ADE}}$ or $\Delta_{\mathrm{IVLA}}$), teacher-forced KL ($\mathrm{KL}_{\mathrm{tf}}$), and hidden-state alignment $\cos(\Delta h,d_{\mathrm{TNAS}})$.
    \item \textbf{Free-generation exact match.} As detailed in \hyperref[app:attack_subset]{Appendices~\ref*{app:attack_subset}}--\ref{app:free_generation_exact}, under free generation, the model autoregressively generates its own text and decodes the corresponding action, and \emph{Exact} reports the fraction of examples whose generated rationale or response matches the clean one word for word.
    \item \textbf{Silent collision rate.} To evaluate whether trajectory divergence creates concrete driving hazards, we score Alpamayo's free-generated $6.4$ s ($64$-waypoint) plans against time-aligned logged traffic actors on the $154$ clips with obstacle annotations~\citep{hu2023uniad,jiang2023vad}, following the protocol detailed in \hyperref[app:open_loop_safety]{Appendix~\ref*{app:open_loop_safety}}.
    At each $0.1$ s planning step, we project the ego vehicle and logged actors as oriented bounding boxes in bird's-eye view, inferring ego heading from consecutive waypoints, and test for overlap using the separating-axis theorem.
    A clip has an open-loop collision if the ego box intersects any actor box at least once over the $6.4$ s horizon, and the \emph{silent collision rate} reports the fraction of the $154$ clips where a collision-free clean plan is perturbed into a collision while the free-generated rationale remains word-for-word identical to the clean rationale.
\end{itemize}

\paragraph{Baselines.}
We compare against two baseline objectives that represent the intuitive output-level approaches to inducing action deviation alone and to inducing action deviation while preserving the generated text.
\mbox{}\\
\textbf{Action-only PGD} only maximizes an action-displacement loss $\mathcal{L}_{\mathrm{act}}(\delta_{\mathrm{img}})$ between the perturbed and clean action predictions over the $T$-step horizon:
\begin{equation}
    \mathcal{L}_{\mathrm{act}}(\delta_{\mathrm{img}})
    =
    \frac{1}{T}\sum_{t=1}^{T}
    D_{\mathrm{act}}\!\bigl(a_t(\delta_{\mathrm{img}}),\,a_t(0)\bigr).
    \label{eq:loss_act}
\end{equation}
Following \hyperref[eq:action_shift]{Equation~\ref*{eq:action_shift}}, the per-step discrepancy $D_{\mathrm{act}}$ is the planar waypoint displacement $\lVert\tau_t(\delta_{\mathrm{img}})-\tau_t(0)\rVert_2$ for Alpamayo and the mean squared coordinate difference across the 7-DoF action vector for InstructVLA.
\mbox{}\\
\textbf{Text-preserving PGD}, detailed in \hyperref[app:input_protocols]{Appendix~\ref*{app:input_protocols}}, subtracts an explicit output-level penalty on the two active text losses defined in \hyperref[sec:method_textspan]{Section~\ref*{sec:method_textspan}}, namely the clean-target NLL increase $\Delta\mathcal{L}_{\mathrm{NLL}}$ and the target-versus-runner-up margin drop $\Delta m$, following joint action--text formulations for reasoning VLAs~\citep{huang2026trap}:
\begin{equation}
    \mathcal{L}_{\mathrm{act+txt}}(\delta_{\mathrm{img}})
    =
    \mathcal{L}_{\mathrm{act}}(\delta_{\mathrm{img}})
    -
    \lambda_{\mathrm{txt}}\,[\Delta\mathcal{L}_{\mathrm{NLL}}+\Delta m]
    \label{eq:loss_act_txt}
\end{equation}

\subsection{TNAS-guided full-image attack}
\label{sec:digital_attack}
\label{sec:attack_baselines}

\begin{table}[t]
\centering
\caption{\textbf{Alpamayo input attacks.} Fixed-prefix and free-generation columns use the canonical $200$ clips. The open-loop column uses the $154$ of them with obstacle annotations, as described in \hyperref[sec:attack_protocol]{Section~\ref*{sec:attack_protocol}}.}
\label{tab:attack_p4_paired}
\label{tab:plan_conflict}
\small
\setlength{\tabcolsep}{4pt}
\begin{tabular}{@{}lrrrrr@{}}
\toprule
 & \multicolumn{3}{c}{Fixed-prefix} & Free gen. & Open loop \\
\cmidrule(lr){2-4}\cmidrule(lr){5-5}\cmidrule(l){6-6}
Attack & \begin{tabular}[c]{@{}r@{}}ADE\\(m) $\uparrow$\end{tabular} & $\mathrm{KL}_{\mathrm{tf}}$ $\downarrow$ & \begin{tabular}[c]{@{}r@{}}TNAS\\cos\end{tabular} & \begin{tabular}[c]{@{}r@{}}Exact\\(\%) $\uparrow$\end{tabular} & \begin{tabular}[c]{@{}r@{}}Silent Coll.\\(\%) $\uparrow$\end{tabular} \\
\midrule
Action-only PGD     & $70.1$ & $2.78\times10^{-1}$ & $4.40\times10^{-4}$ & $9.5$ & $4.55$ \\
Text-preserving PGD & $69.2$ & $3.26\times10^{-1}$ & $8.00\times10^{-4}$ & $83.0$ & $48.7$ \\
\midrule
\multicolumn{6}{@{}l}{TNAS Full image} \\
$\lambda_{\mathrm{cos}}=5$  & $67.4$ & $3.32\times10^{-1}$ & $1.64\times10^{-2}$ & $80.5$ & $46.8$ \\
$\lambda_{\mathrm{cos}}=10$ & $67.8$ & $2.99\times10^{-1}$ & $3.93\times10^{-2}$ & $81.0$ & $47.4$ \\
$\lambda_{\mathrm{cos}}=15$ & $65.5$ & $2.46\times10^{-1}$ & $7.57\times10^{-2}$ & $84.0$ & $45.5$ \\
$\lambda_{\mathrm{cos}}=20$ & $63.2$ & $1.89\times10^{-1}$ & $1.33\times10^{-1}$ & $88.0$ & $45.5$ \\
$\lambda_{\mathrm{cos}}=30$ & $53.0$ & $\underline{1.03\times10^{-1}}$ & $\underline{2.46\times10^{-1}}$ & $\underline{89.5}$ & $\underline{52.6}$ \\
\midrule
\multicolumn{6}{@{}l}{TNAS Patch} \\
$\lambda_{\mathrm{cos}}=5$  & $38.7$ & $1.39\times10^{-1}$ & $1.55\times10^{-2}$ & $70.5$ & $33.8$ \\
$\lambda_{\mathrm{cos}}=10$ & $39.8$ & $1.24\times10^{-1}$ & $4.24\times10^{-2}$ & $70.5$ & $\underline{35.1}$ \\
$\lambda_{\mathrm{cos}}=15$ & $37.2$ & $1.07\times10^{-1}$ & $6.74\times10^{-2}$ & $71.0$ & $31.8$ \\
$\lambda_{\mathrm{cos}}=20$ & $34.8$ & $9.00\times10^{-2}$ & $1.04\times10^{-1}$ & $76.0$ & $34.4$ \\
$\lambda_{\mathrm{cos}}=30$ & $30.0$ & $\underline{5.70\times10^{-2}}$ & $\underline{1.62\times10^{-1}}$ & $\underline{80.5}$ & $34.4$ \\
\bottomrule
\end{tabular}
\end{table}

\paragraph{Output-level action and text objectives compete.}
In \hyperref[tab:attack_p4_paired]{Table~\ref*{tab:attack_p4_paired}}, Action-only PGD in Alpamayo shifts the planned trajectory by $70.1$ m ADE, but it also rewrites the driving rationale on $90.5\%$ of clips, leaving only $4.55\%$ silent collisions.
Adding the output-level text penalty in Text-preserving PGD keeps the clean rationale intact on $83.0\%$ of clips and raises silent collisions to $48.7\%$, yet the two objectives work against each other during optimization.
In \hyperref[eq:loss_act_txt]{Equation~\ref*{eq:loss_act_txt}}, both terms update the same input pixels: the action loss maximizes trajectory displacement, while the text penalty constrains only the clean target tokens' NLL and top-1 margin.
Holding those target tokens in place against unguided action updates pushes vocabulary-wide teacher-forced KL from $2.78\times10^{-1}$ under Action-only PGD to $3.26\times10^{-1}$, which lowers the chance that stochastic decoding reproduces the clean rationale.

\paragraph{Resolving the conflict with TNAS guidance.}
To resolve this conflict at its geometric source, \textbf{TNAS-guided PGD} augments the text-preserving objective with a term that rewards cosine alignment between the input-induced hidden-state shift $\Delta h(\delta_{\mathrm{img}})$ and each example's TNAS direction $d_{\mathrm{TNAS}}$:
\begin{equation}
    \mathcal{L}_{\mathrm{TNAS}}(\delta_{\mathrm{img}})
    =
    \mathcal{L}_{\mathrm{act+txt}}(\delta_{\mathrm{img}})
    +
    \lambda_{\mathrm{cos}}\,
    \cos\bigl(\Delta h(\delta_{\mathrm{img}}),d_{\mathrm{TNAS}}\bigr).
    \label{eq:loss_tnas}
\end{equation}
As shown in \hyperref[sec:mechanism]{Section~\ref*{sec:mechanism}}, shifting the hidden state along $d_{\mathrm{TNAS}}$ drives action displacement while leaving the two text losses in \hyperref[eq:loss_act_txt]{Equation~\ref*{eq:loss_act_txt}} unchanged to first order.
Maximizing cosine similarity in \hyperref[eq:loss_tnas]{Equation~\ref*{eq:loss_tnas}} directly steers the direction of $\Delta h(\delta_{\mathrm{img}})$ toward $d_{\mathrm{TNAS}}$.
During optimization, this allows PGD updates to increase action displacement with minimal disturbance to the text logits and reduces the conflict between the action and text terms in $\mathcal{L}_{\mathrm{act+txt}}$.

\paragraph{TNAS guidance quiets the text channel.}
In \hyperref[tab:attack_p4_paired]{Table~\ref*{tab:attack_p4_paired}}, increasing $\lambda_{\mathrm{cos}}$ from $5$ to $30$ raises cosine alignment between $\Delta h$ and $d_{\mathrm{TNAS}}$ at every step from $1.64\times10^{-2}$ to $2.46\times10^{-1}$, compared with $8.00\times10^{-4}$ under Text-preserving PGD.
Even though the cosine term never accesses token logits or text losses, steering the hidden state toward $d_{\mathrm{TNAS}}$ pulls teacher-forced KL down monotonically across the sweep to $1.03\times10^{-1}$ at $\lambda_{\mathrm{cos}}=30$, a $68.4\%$ reduction relative to Text-preserving PGD.
Because lower distributional drift keeps the perturbed next-token distribution closer to the clean one during stochastic decoding, free-generation exact match also rises monotonically across the sweep and reaches $89.5\%$ at $\lambda_{\mathrm{cos}}=30$, up from $83.0\%$ without TNAS guidance.

\paragraph{Steering toward $d_{\mathrm{TNAS}}$ concentrates displacement into silent collisions.}
Under a fixed pixel budget, weighting the objective toward $d_{\mathrm{TNAS}}$ trades redundant fixed-prefix displacement, from $69.2$ m under Text-preserving PGD to $53.0$ m at $\lambda_{\mathrm{cos}}=30$, for stronger TNAS alignment without reducing collisions.
On the $154$ obstacle-annotated scenes, the free-generated plans continue to collide on $61.7\%$ of clips at $\lambda_{\mathrm{cos}}=30$ versus $61.0\%$ under Text-preserving PGD, so the higher exact rationale preservation translates directly into silent collisions.
The final step of the sweep brings the largest gains in text stability. Between $\lambda_{\mathrm{cos}}=20$ and $\lambda_{\mathrm{cos}}=30$, TNAS alignment nearly doubles and teacher-forced KL falls the most, while silent collisions reach $52.6\%$, the highest in the sweep.
TNAS guidance therefore buys text stability without giving up silent collisions.

\begin{table}[t]
\centering
\caption{\textbf{InstructVLA input attacks.} Fixed-prefix and free-generation columns use the canonical $200$ examples.}
\label{tab:ivla_pixel}
\label{tab:ivla_paired}
\small
\begin{tabular}{@{}lrrrr@{}}
\toprule
 & \multicolumn{3}{c}{Fixed-prefix} & Free gen. \\
\cmidrule(lr){2-4}\cmidrule(l){5-5}
Attack & \begin{tabular}[c]{@{}r@{}}Action shift\\(norm.) $\uparrow$\end{tabular} & $\mathrm{KL}_{\mathrm{tf}}$ $\downarrow$ & \begin{tabular}[c]{@{}r@{}}TNAS\\cos\end{tabular} & \begin{tabular}[c]{@{}r@{}}Exact\\(\%) $\uparrow$\end{tabular} \\
\midrule
Action-only PGD     & $2.04$ & $4.66\times10^{-2}$ & $1.20\times10^{-3}$ & $72.5$ \\
Text-preserving PGD & $1.97$ & $3.46\times10^{-2}$ & $2.90\times10^{-3}$ & $77.5$ \\
\midrule
\multicolumn{5}{@{}l}{TNAS Full image} \\
$\lambda_{\mathrm{cos}}=0.10$ & $1.85$ & $2.67\times10^{-2}$ & $1.22\times10^{-1}$ & $81.0$ \\
$\lambda_{\mathrm{cos}}=0.15$ & $1.80$ & $2.35\times10^{-2}$ & $1.66\times10^{-1}$ & $81.5$ \\
$\lambda_{\mathrm{cos}}=0.20$ & $1.73$ & $\underline{2.06\times10^{-2}}$ & $\underline{2.13\times10^{-1}}$ & $\underline{85.0}$ \\
\midrule
\multicolumn{5}{@{}l}{TNAS Patch} \\
$\lambda_{\mathrm{cos}}=0.20$ & $1.32$ & $2.80\times10^{-2}$ & $1.13\times10^{-1}$ & $77.5$ \\
\bottomrule
\end{tabular}
\end{table}

\paragraph{Replication in InstructVLA.}
\label{sec:ivla_pixel}
In \hyperref[tab:ivla_pixel]{Table~\ref*{tab:ivla_pixel}}, the same geometric guidance transfers to robotic manipulation in InstructVLA, where perturbations modify only the VLM visual stream while a single task-intention sentence averaging $9.60$ words, such as ``I will move the cup to the sink,'' covers an entire $16$-step 7-DoF action chunk.
Across the $\lambda_{\mathrm{cos}}$ sweep from $0.10$ to $0.20$, cosine alignment between $\Delta h$ and $d_{\mathrm{TNAS}}$ rises at every step from $1.22\times10^{-1}$ to $2.13\times10^{-1}$, whereas Action-only PGD at $1.20\times10^{-3}$ and Text-preserving PGD at $2.90\times10^{-3}$ leave the hidden-state shift unaligned with $d_{\mathrm{TNAS}}$.
Even the smallest guided weight $\lambda_{\mathrm{cos}}=0.10$ already improves over Text-preserving PGD on every text metric, and steering further toward $d_{\mathrm{TNAS}}$ pulls teacher-forced KL down monotonically from $2.67\times10^{-2}$ to $2.06\times10^{-2}$ at $\lambda_{\mathrm{cos}}=0.20$, a $40.5\%$ reduction from $3.46\times10^{-2}$ under Text-preserving PGD and $55.8\%$ below $4.66\times10^{-2}$ under Action-only PGD.
This lower distributional drift raises free-generation exact match monotonically across the sweep from $81.0\%$ to $85.0\%$ at $\lambda_{\mathrm{cos}}=0.20$, up from $77.5\%$ under Text-preserving PGD.
Even though this coarser many-to-one intention bottleneck already leaves $72.5\%$ of responses unchanged under Action-only PGD without any text penalty, steering along $d_{\mathrm{TNAS}}$ still eliminates one-third of the response changes that remain under Text-preserving PGD.
Under the fixed pixel budget, weighting the objective toward $d_{\mathrm{TNAS}}$ trades only a small part of the fixed-prefix action shift, from $2.04$ under Action-only PGD and $1.97$ under Text-preserving PGD to $1.73$ at $\lambda_{\mathrm{cos}}=0.20$, so it retains $87.5\%$ of Text-preserving PGD's action displacement alongside this stronger text stability.
As in Alpamayo, the final step from $\lambda_{\mathrm{cos}}=0.15$ to $\lambda_{\mathrm{cos}}=0.20$ brings the largest exact-match gain within the guided sweep, jumping from $81.5\%$ to $85.0\%$ as TNAS alignment climbs from $1.66\times10^{-1}$ to $2.13\times10^{-1}$ and teacher-forced KL drops from $2.35\times10^{-2}$ to $2.06\times10^{-2}$.

\takeaway{\textbf{TNAS guidance resolves the action--text optimization conflict, suppressing text drift while increasing silent collisions.} Output-level text penalties compete with the action gradient and leave hidden-state updates unaligned with $d_{\mathrm{TNAS}}$. Steering along $d_{\mathrm{TNAS}}$ lowers teacher-forced KL by $68.4\%$ in Alpamayo and $40.5\%$ in InstructVLA, raises exact rationale preservation from $83.0\%$ to $89.5\%$ and from $77.5\%$ to $85.0\%$, and produces silent collisions on $52.6\%$ of annotated driving scenes alongside a $1.73$-unit manipulation action shift.}

\subsection{Localized patch attack}
\label{sec:patch_attack}

Our full-image attacks assume a digital attacker who can add a perturbation to the preprocessed image tensor fed to the vision encoder, which in practice requires access to the model's software pipeline.
Moving toward localized attacks on objects in the scene, such as a patch on the back of a front vehicle, requires the perturbation to operate within a restricted spatial region while remaining consistent across views and time.
To test whether an attacker can still steer the shared representation along $d_{\mathrm{TNAS}}$ under these constraints, we confine the perturbation to a localized patch and retain the TNAS-guided objective.

\paragraph{Shared forward-camera patch in Alpamayo.}
In Alpamayo, we optimize a single clip-specific patch on the road surface ahead of the vehicle and tie it across the two forward-facing cameras and all four timestamps, covering $13.1\%$ of visual tokens.
The front-telephoto crop receives the base pattern, the front-wide crop receives a bilinearly resized version on the corresponding road region, and the two cross-view cameras remain clean.
In \hyperref[tab:attack_p4_paired]{Table~\ref*{tab:attack_p4_paired}}, increasing $\lambda_{\mathrm{cos}}$ from $5$ to $30$ raises cosine alignment between $\Delta h$ and $d_{\mathrm{TNAS}}$ at every step from $1.55\times10^{-2}$ to $1.62\times10^{-1}$ and pulls teacher-forced KL down monotonically from $1.39\times10^{-1}$ to $5.70\times10^{-2}$, a $59.0\%$ reduction that lifts free-generation exact match from $70.5\%$ to $80.5\%$.
With the surrounding road and cross-view cameras left untouched, fixed-prefix trajectory displacement reaches $38.7$ m ADE at $\lambda_{\mathrm{cos}}=5$ and $39.8$ m at $\lambda_{\mathrm{cos}}=10$, and weighting the objective toward $\lambda_{\mathrm{cos}}=30$ trades part of this shift to $30.0$ m for stronger text stability while keeping silent collisions between $31.8\%$ and $35.1\%$ of obstacle-annotated scenes.

\paragraph{VLM-branch patch in InstructVLA.}
In InstructVLA, we confine SigLIP updates to a $0.4\times 0.4$ lower-center rectangle covering $16\%$ of the image area.
In \hyperref[tab:ivla_pixel]{Table~\ref*{tab:ivla_pixel}}, TNAS-guided PGD at $\lambda_{\mathrm{cos}}=0.20$ steers the hidden-state shift from this patch to $1.13\times10^{-1}$ cosine alignment with $d_{\mathrm{TNAS}}$ and lowers teacher-forced KL to $2.80\times10^{-2}$, $19.1\%$ below the $3.46\times10^{-2}$ of full-image Text-preserving PGD.
Free-generation exact match reaches $77.5\%$, matching full-image Text-preserving PGD and exceeding the $72.5\%$ of full-image Action-only PGD.
With the remaining $84\%$ of the image left untouched, the patch shifts the predicted action chunk by $1.32$ and retains $76.4\%$ of the $1.73$ displacement from full-image TNAS-guided PGD at $\lambda_{\mathrm{cos}}=0.20$.

\takeaway{\textbf{TNAS-guided action divergence persists under localized patch constraints.} In Alpamayo, a forward-camera patch on $13.1\%$ of visual tokens cuts teacher-forced KL by $59.0\%$ across the $\lambda_{\mathrm{cos}}$ sweep, preserves up to $80.5\%$ of rationales word for word, and induces silent collisions on up to $35.1\%$ of annotated scenes. In InstructVLA, a patch on $16\%$ of the image shifts the action chunk by $1.32$ units while matching the $77.5\%$ exact preservation of full-image Text-preserving PGD with $19.1\%$ lower teacher-forced KL.}

\section{Related Work}
\label{sec:related_work}

\paragraph{Reasoning vision-language-action models.}
VLA policies connect visual and language representations to embodied control~\citep{brohan2022rt,zitkovich2023rt,driess2023palm,kim2024openvla}, with autonomous driving models and benchmarks extending this paradigm to multi-camera road scenes~\citep{zhou2025autovla,arai2025covla,zhou2026opendrivevla,fu2025orion,renz2025simlingo,shao2023lmdrive,li2025recogdrive}.
Reasoning-conditioned VLAs expose natural-language scene descriptions, plans, or chain-of-thought traces alongside continuous action prediction~\citep{tian2024drivevlm,jiang2024senna,yin2025deepthinkvla,zawalski2024robotic}.
We study Alpamayo-1.5~\citep{wang2025alpamayo} and InstructVLA~\citep{yang2025instructvla}, two representative embodiments that pair autoregressive VLM text generation with continuous diffusion or flow-matching action heads reading from a shared hidden state.

\paragraph{Monitoring model reasoning.}
Chain-of-thought monitoring treats verbalized reasoning as a safety signal and asks when that signal stops reflecting behavior~\citep{baker2025monitoring,korbak2025monitorability}.
In language models, faithfulness studies distinguish plausible post-hoc explanations from the causal drivers of an answer~\citep{turpin2023language,lanham2023measuring,arcuschin2025chainofthought,lyu-etal-2023-faithful}.
In embodied VLAs, prior work addresses unperturbed mismatches between textual plans and actions at the output level through RL consistency rewards~\citep{wang2025alpamayo} or runtime action verification~\citep{wu2026say}.
Concurrent with our work, \citet{foutter2026faithfulness} observe that Alpamayo-1.5 can state that it will yield while steering without decelerating, and post-train driving planners with a learned critic that rewards semantic consistency among observations, reasoning steps, and trajectories.
We instead diagnose why text and action decouple at the shared hidden state, where TNAS shows that the locally most action-sensitive direction is nearly orthogonal to the active text-loss gradients, and steer visual inputs along this subspace to alter the physical action while keeping the generated rationale word-for-word identical.
The TNAS geometry justifies pairing rationale monitors with execution-channel monitors~\citep{seligmann2026vlafail}.

\paragraph{Adversarial attacks on driving and embodied policies.}
Adversarial examples were first studied as imperceptible $\ell_p$-bounded perturbations of the full digital image~\citep{szegedy2014intriguing,goodfellow2015explaining,carlini2017towards,madry2018towards}.
Later work showed that such perturbations survive printing, camera capture, and viewpoint changes~\citep{kurakin2017adversarial,athalye2018synthesizing}, and that a localized patch such as a printed sticker or accessory can fool image classifiers and face recognition systems~\citep{brown2017adversarial,karmon2018lavan,sharif2016accessorize,eykholt2018robust}.
In autonomous driving, physical patches appear as dirt-like road patterns that steer automated lane centering out of its lane~\citep{sato2021dirty}, roadside billboards that mislead steering models~\citep{kong2020physgan}, and patches on the rear of a lead vehicle that corrupt monocular depth estimation~\citep{cheng2022physical}.
Attacks on driving stacks also target trajectory prediction~\citep{zhang2022adversarial} and collaborative perception~\citep{zhang2024data}.
On VLAs, a small patch in the camera view can derail robotic manipulation~\citep{wang2025exploring} or hijack chain-of-thought reasoning~\citep{huang2026trap}, and other attacks corrupt the reasoning trace directly~\citep{trinh2026altered} or implant backdoors~\citep{li2025attackvla,zhou2025badvla,li2026vlasafety}.
Reasoning-hijacking attacks visibly alter the text, and unconstrained action attacks corrupt it as a byproduct.
Our TNAS-guided attack instead holds the generated rationale fixed while steering the action, under both full-image perturbations and a localized patch.

\paragraph{Representation steering and subspace projection.}
Representation engineering and activation steering change model behavior by adding a direction to intermediate hidden states at inference time~\citep{zou2023representation,turner2025steering}.
Linear concept erasure projects hidden states so that a linear readout can no longer recover a protected attribute~\citep{ravfogel2020null,belrose2023leace}.
TNAS applies this projection to the steering direction itself at the shared text--action interface.
Removing the active text-loss span from the locally most action-sensitive direction leaves a residual that moves the action readout while keeping the text readout unchanged to first order.
Activation steering writes such directions into the hidden state, whereas our attack reaches the TNAS residual by perturbing only the visual input.

\section{Conclusion}
\label{sec:conclusion}

We studied rationale-preserving action divergence in reasoning VLAs, where a perturbed visual input alters the physical action while leaving the generated rationale word for word unchanged.
To diagnose this vulnerability at the shared text--action interface, we introduced the text-neutral action-subspace (TNAS) probe, which projects active text-loss gradients out of the locally most action-sensitive hidden-state direction to isolate a text-neutral action residual.
Across Alpamayo for autonomous driving and InstructVLA for robotic manipulation, the action-sensitive direction overlaps the active text span by only $1.87\%$ and $5.28\%$ of its norm, and equal-norm interventions along the TNAS residual shift actions $4.64\times$ to $47.8\times$ more than text-gradient perturbations while text-gradient perturbations induce $21.2\times$ to $277\times$ more text drift.
Steering visual perturbations along this residual resolves the optimization conflict between action displacement and text preservation and cuts teacher-forced text drift by $68.4\%$ in Alpamayo and $40.5\%$ in InstructVLA relative to text-preserving baselines.
Under free generation, TNAS-guided attacks preserve the rationale word for word while inducing new collisions on $52.6\%$ of annotated driving scenes under full-image perturbations and $35.1\%$ under a localized forward-camera patch, and preserve $85.0\%$ of manipulation responses alongside a $1.73$-unit action-chunk shift.
Because an unchanged rationale cannot certify a safe action, embodied safety monitors must verify the predicted action channel directly alongside the text.

\begin{ack}
We thank the area chair and the anonymous reviewers for their valuable feedback and constructive suggestions. We also specifically thank Chufan Shi for his valuable help and insightful discussions throughout this work. This material is based upon work supported by the National Science Foundation under Grant No. 2333489. The authors declare no competing interests.
\end{ack}

{\small
\bibliographystyle{unsrtnat}
\bibliography{references}
}

\appendix
\newpage
\section{Appendix}

The appendix gives the protocols and full values behind the main-text claims. The cohorts that every experiment shares are fixed in \hyperref[app:dataset]{Appendix~\ref*{app:dataset}}. \hyperref[app:input_protocols]{Appendix~\ref*{app:input_protocols}} specifies the attacks, metrics, and paired tests behind \hyperref[tab:attack_p4_paired]{Tables~\ref*{tab:attack_p4_paired}} and~\ref{tab:ivla_pixel}, and \hyperref[app:open_loop_safety]{Appendix~\ref*{app:open_loop_safety}} defines the silent-collision readout. The TNAS construction and the hook choice of \hyperref[sec:setup]{Section~\ref*{sec:setup}} are detailed in \hyperref[app:local_probe_impl]{Appendix~\ref*{app:local_probe_impl}}. For \hyperref[sec:geometry]{Probe~1}, \hyperref[app:benign_reader]{Appendix~\ref*{app:benign_reader}} evaluates the rationale-only reader and \hyperref[app:projection]{Appendix~\ref*{app:projection}} audits the projections. \hyperref[app:probe_tables]{Appendix~\ref*{app:probe_tables}} lists the values plotted in \hyperref[fig:probes]{Figure~\ref*{fig:probes}}, and \hyperref[app:sign_aligned]{Appendix~\ref*{app:sign_aligned}} details the sign-aligned amplification protocol of \hyperref[sec:nulling_and_amplification]{Probe~3}. Asset terms and compute resources close the appendix in \hyperref[app:asset_licenses]{Appendices~\ref*{app:asset_licenses}} and~\ref{app:compute_resources}.

\subsection{Datasets and paired cohorts}
\label{app:dataset}

\paragraph{Alpamayo evaluation subset.}
The Alpamayo experiments use the gated NVIDIA \texttt{PhysicalAI-\allowbreak Autonomous-\allowbreak Vehicles} HuggingFace dataset together with the \texttt{nvidia/\allowbreak Alpamayo-1.5-10B} checkpoint. We start from the clip manifest distributed with the model release, which lists $1{,}181$ unique clip IDs, and download each clip's egomotion stream and four camera streams through the \texttt{physical\_ai\_av} dataset interface: the $120$\textdegree{} cross-left, front-wide, and cross-right cameras and the $30$\textdegree{} front-telephoto camera. Probes read each clip at the canonical reference time $t_0=5{,}100{,}000~\mu\mathrm{s}$ with all four cameras and four temporal frames, for $16$ images per clip.

The 200-clip subset is constructed deterministically. We list every clip ID that has a non-empty clean rationale and a resolved local cache, sort the list lexicographically, shuffle it with Python's \texttt{random.Random(0)}, and take the first $200$ entries. All Alpamayo probe and input experiments draw from this one ordered list.

\paragraph{InstructVLA evaluation subset.}
The InstructVLA experiments use $200$ Bridge-distribution examples from the VLA-IT language evaluation set. The released VLA-IT split fixes the example list, so no sampling seed is involved. The hidden-state probes and the full-image and patch input evaluations all use this ordered list, and every comparison is paired by example ID. \hyperref[app:layer]{Appendices~\ref*{app:layer}} and~\ref{app:ivla_impl} describe hook selection and the differentiable action readout.

\paragraph{Evaluation cohorts.}
\label{app:attack_subset}
Each model's probe and input results use its $200$-example subset. The Alpamayo safety analysis uses the $154$ of those $200$ clips that have obstacle annotations in the dataset release, as described in \hyperref[app:open_loop_safety]{Appendix~\ref*{app:open_loop_safety}}. \hyperref[app:projection]{Appendix~\ref*{app:projection}} also reports the projection ratio on the $1{,}180$ manifest clips with a nonempty rationale, and the benign rationale-only reader of \hyperref[app:benign_reader]{Appendix~\ref*{app:benign_reader}} uses all $1{,}181$ clips of the manifest.
\subsection{Input protocols and text/action metrics}
\label{app:input_protocols}

\paragraph{Alpamayo attack settings.}
All Alpamayo attacks use the objectives of \hyperref[sec:attack_protocol]{Section~\ref*{sec:attack_protocol}} with text weight $\lambda_{\mathrm{txt}}=1$. The text penalty is the raw change in clean-target NLL plus the target-margin drop, with no hinge. During optimization, both trajectories come from a differentiable $4$-step flow rollout with seed $42$, and every reported readout uses the $10$-step solver of \hyperref[app:action_metrics]{Appendix~\ref*{app:action_metrics}}. PGD starts from uniform noise in $[-\alpha,\alpha]$, where $\alpha$ is the step size, drawn with seed $0$ for each clip and method, and every row reports the final step $200$.

\begin{figure}[t]
\centering
\includegraphics[width=\linewidth]{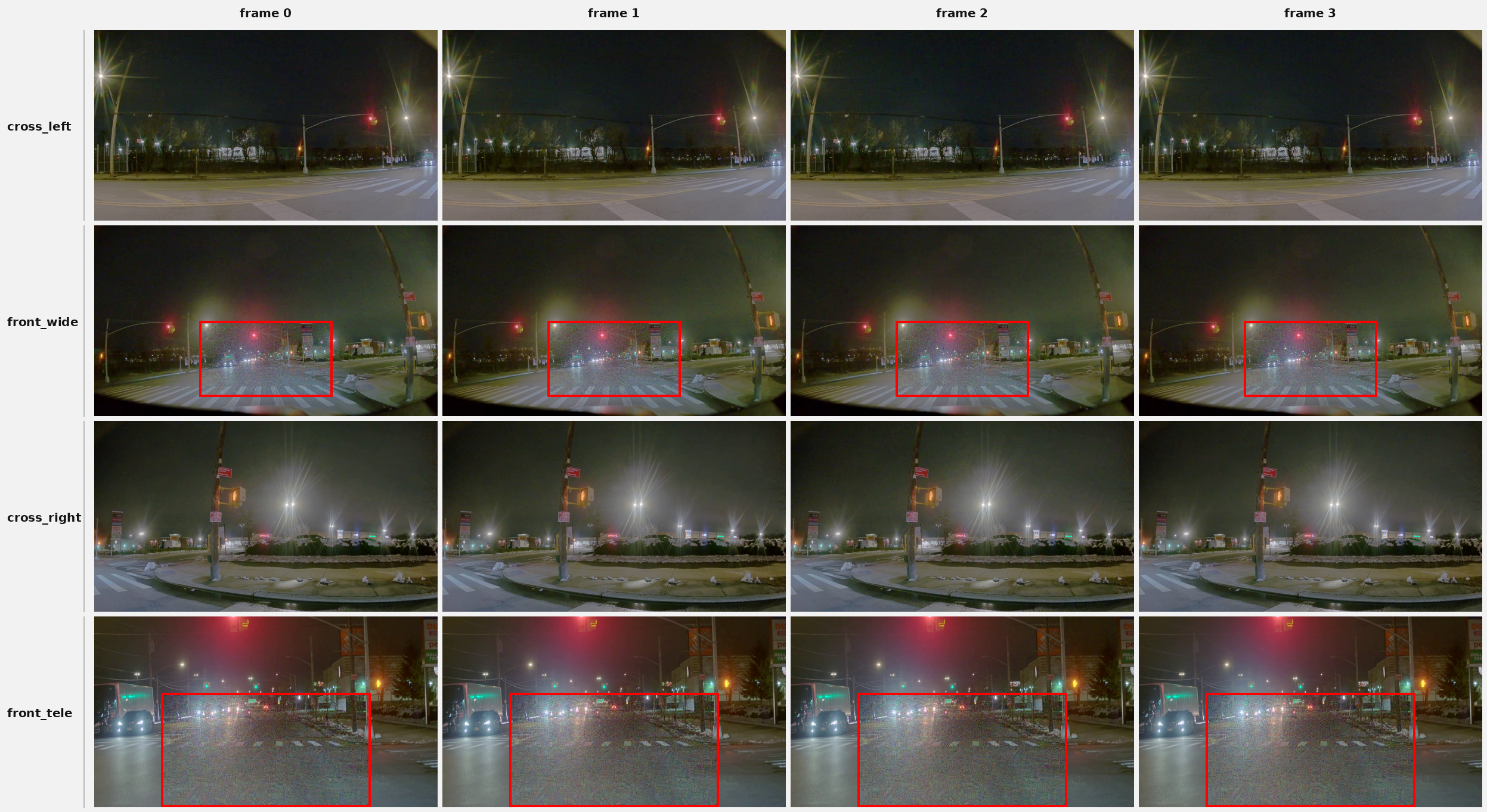}\\[3pt]
\includegraphics[width=\linewidth]{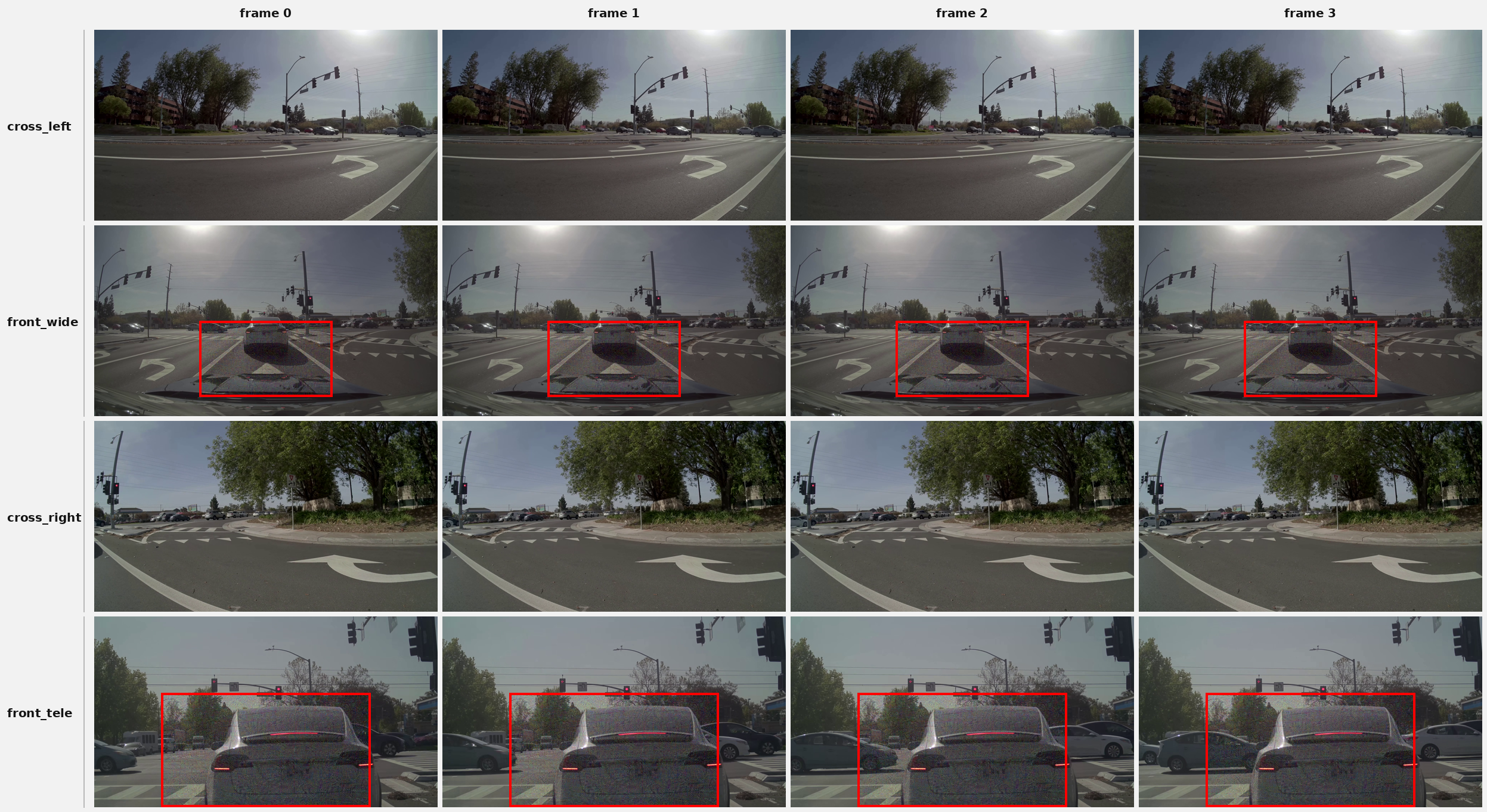}
\caption{\textbf{Shared forward-camera patch in Alpamayo.} Illustrative patches on two clips of the evaluation cohort, optimized for $200$ PGD steps at $\lambda_{\mathrm{cos}}=20$ in the patch layout of \hyperref[tab:attack_p4_paired]{Table~\ref*{tab:attack_p4_paired}}. Rows are the four cameras and columns the four input frames. Red boxes mark the patch. In every frame, the optimized pattern fills the front-telephoto box and a resized copy fills the front-wide box, while the cross-view cameras stay clean.}
\label{fig:patch_demo}
\end{figure}

The patch attack optimizes one pattern per clip. The processor resizes each camera frame to $320\times576$ pixels, a grid of $20\times36$ input patches of $16\times16$ pixels. In the front-telephoto frame, the pattern covers $12\times22$ patches, or $192\times352$ pixels, centered horizontally and flush with the bottom edge. A bilinear resize of the whole pattern covers $8\times14$ patches, or $128\times224$ pixels, of the front-wide frame, centered horizontally with its top edge at the vertical center. Both copies repeat in all four frames of their camera, and the cross-left and cross-right frames stay clean. The patched area is $1{,}504$ of the $11{,}520$ input patches, or $13.1\%$. \hyperref[fig:patch_demo]{Figure~\ref*{fig:patch_demo}} shows this layout on two clips.

\paragraph{InstructVLA text-loss adaptation.}
The InstructVLA text-preserving baseline uses a one-sided hinge on clean-response positions $R$. Let $m_t$ be the target-token logit minus the largest non-target logit, let $\mathcal{L}_{\mathrm{NLL}}$ be mean target-token NLL, and write $[x]_+=\max(x,0)$. Its text penalty is
\[
    \mathcal{L}_{\mathrm{txt}}^{\mathrm{IVLA}}
    =\frac{1}{|R|}\sum_{t\in R}
      [m_t^{\mathrm{cln}}-m_t^{\mathrm{adv}}-\tau_m]_+
     +[\mathcal{L}_{\mathrm{NLL}}^{\mathrm{adv}}
       -\mathcal{L}_{\mathrm{NLL}}^{\mathrm{cln}}-\tau_n]_+ .
\]
The reported hinge family uses $\tau_m=\tau_n=0$, unit action and text weights, and no additional KL penalty. The maximized objective is the mean squared action-coordinate displacement minus this penalty, plus $\lambda_{\mathrm{cos}}\cos(\Delta h,d_{\mathrm{TNAS}})$ for guided rows. The text-preserving baseline is the $\lambda_{\mathrm{cos}}=0$ member, and the action-only arm drops both the text and the guidance terms. Within the full-image hinge family, the runs differ only in the guidance weight.

\paragraph{InstructVLA attack settings.}
The localized run uses a rectangular mask of $40\%$ of the input height and $40\%$ of its width, centered below the image center, and keeps the perturbed input inside $[-1,1]$. As in Alpamayo, PGD starts from seeded uniform noise in $[-\alpha,\alpha]$, restricted to the mask for the patch, and every row reports the final step $200$. The perturbation enters the VLM/SigLIP input, and the action expert keeps its clean DINO input in both full-image and patch runs.

\subsubsection{Action decoding and metrics}
\label{app:action_metrics}

Both models decode actions with an explicit $10$-step Euler flow-matching solver from $t=0$ to $t=1$, without classifier-free guidance, and share the initial Gaussian noise between paired runs through seed $42$. Alpamayo's solver state is a $64\times2$ sequence of normalized acceleration and curvature controls, which the released unicycle action space integrates into the waypoints of \hyperref[eq:action_shift]{Equation~\ref*{eq:action_shift}}. InstructVLA's solver state is the $16\times7$ normalized action chunk itself, and $\Delta_{\mathrm{IVLA}}$ reads the sampler's normalized outputs before deployment clipping, gripper thresholding, and dataset-specific unnormalization, so its coordinates can leave $[-1,1]$.
\subsubsection{Definition and implementation of teacher-forced KL}
\label{app:tf_kl}

The teacher-forced KL metric of \hyperref[sec:metrics]{Section~\ref*{sec:metrics}} is the forward KL from the clean to the perturbed next-token distribution, summed over the vocabulary and averaged over the active rationale/response span. With clean and perturbed per-token logits $z^{\mathrm{cln}}_t,z^{\mathrm{adv}}_t\in\mathbb{R}^V$ and $p^{\bullet}_t=\mathrm{softmax}(z^{\bullet}_t)$,
\begin{equation}
    \mathrm{KL}_{\mathrm{tf}}(\delta) \;=\; \frac{1}{|R|}\sum_{t\in R}\sum_{v=1}^{V}
    p^{\mathrm{cln}}_{t,v}\bigl[\log p^{\mathrm{cln}}_{t,v} - \log p^{\mathrm{adv}}_{t,v}\bigr],
    \label{eq:tf_kl_appendix}
\end{equation}
which is computed in log-space for numerical stability by applying \texttt{log\_softmax} separately to $z^{\mathrm{cln}}_t$ and $z^{\mathrm{adv}}_t$ before forming the inner sum. The metric is reported in nats. The clean rationale or response is fixed as the teacher-forced target sequence at every $t\in R$. At each position, the paired runs share the textual prefix and target token, and they differ only through the specified hidden-state intervention for local probes or visual-input intervention for input attacks.

\paragraph{Cross-table reading.} Hidden-state probes select steps from finite alpha grids under the model-specific KL-budget rules of \hyperref[app:local_probe_impl]{Appendix~\ref*{app:local_probe_impl}}, while input attacks optimize under visual-input budgets with no fixed KL target. Both report \hyperref[eq:tf_kl_appendix]{Equation~\ref*{eq:tf_kl_appendix}} on the corresponding model's active text span.

\subsubsection{Free-generation exact match}
\label{app:free_generation_exact}

\paragraph{Alpamayo sampled decoding.}
For each clean or perturbed visual input, we run the VLM autoregressively without teacher forcing, with one sampled sequence, top-$p=0.98$, temperature $0.6$, at most $256$ new tokens, and generation seed $42$ shared between paired runs. Generation stops after the trajectory-start token, after which we extract the rendered chain-of-thought span. Sharing the random seed between paired clean and perturbed runs isolates the effect of the visual perturbation from sampling noise.

\paragraph{InstructVLA greedy decoding.}
Clean and perturbed responses are generated separately with sampling disabled and a single beam. The decoder extracts the generated suffix, truncates it at the designated meta and end markers, \texttt{<new\_token\_0>} and \texttt{<|im\_end|>}, and strips surrounding whitespace. Exact match is equality of these cleaned strings. The $200$ clean responses are short, templated intention statements of $9.60$ words on average, with a median of $9$, a range of $5$ to $17$, and $178$ distinct strings. Each one summarizes a full $16$-step action chunk, the many-to-one intention bottleneck of \hyperref[sec:ivla_pixel]{Section~\ref*{sec:ivla_pixel}}.

For either model, if $c_{\mathrm{cln}}$ and $c_{\mathrm{adv}}$ are the strings extracted under its protocol, the metric is
\begin{equation}
    \mathrm{Exact}_{\mathrm{free}} = \mathbf{1}\{c_{\mathrm{cln}} = c_{\mathrm{adv}}\}.
\end{equation}

\subsubsection{Paired tests for the input tables}
\label{app:input_stats}

Every row of \hyperref[tab:attack_p4_paired]{Tables~\ref*{tab:attack_p4_paired}} and~\ref{tab:ivla_pixel} is paired by example through the shared seeds above, so \hyperref[tab:input_paired]{Table~\ref*{tab:input_paired}} compares arms with exact paired tests on the comparisons interpreted in \hyperref[sec:attack]{Section~\ref*{sec:attack}}. At the strongest guidance weights, TNAS guidance significantly lowers teacher-forced KL and raises exact match over Text-preserving PGD in both models. In Alpamayo, its attacked-plan and silent-collision counts stay level with Text-preserving PGD and far above Action-only PGD. The InstructVLA patch matches the exact match of full-image Text-preserving PGD with significantly lower KL.

\begin{table}[t]
\centering
\caption{\textbf{Paired tests for the input tables.} Binary outcomes list the counts of the two arms and the discordant examples favoring each arm, tested with exact McNemar tests. KL rows count the examples on which the first arm has lower teacher-forced KL, tested with exact sign tests. Exact match and KL use $200$ examples, and collisions use the $154$ annotated clips.}
\label{tab:input_paired}
\small
\begin{tabular}{@{}lrrr@{}}
\toprule
Outcome & Counts & Discordant & $p$ \\
\midrule
\multicolumn{4}{@{}l}{Alpamayo: full image $\lambda_{\mathrm{cos}}=30$ vs.\ Text-preserving PGD} \\
Exact match & $179$ vs.\ $166$ & $20$/$7$ & $0.0192$ \\
Lower $\mathrm{KL}_{\mathrm{tf}}$ & $179$ of $200$ & --- & $1.95\times10^{-32}$ \\
Attacked collision & $95$ vs.\ $94$ & $22$/$21$ & $1.00$ \\
Silent collision & $81$ vs.\ $75$ & $25$/$19$ & $0.451$ \\
\midrule
\multicolumn{4}{@{}l}{Alpamayo: full image $\lambda_{\mathrm{cos}}=30$ vs.\ Action-only PGD} \\
Silent collision & $81$ vs.\ $7$ & $75$/$1$ & $2.04\times10^{-21}$ \\
\midrule
\multicolumn{4}{@{}l}{InstructVLA: full image $\lambda_{\mathrm{cos}}=0.20$ vs.\ Text-preserving PGD} \\
Exact match & $170$ vs.\ $155$ & $24$/$9$ & $0.0135$ \\
Lower $\mathrm{KL}_{\mathrm{tf}}$ & $163$ of $200$ & --- & $4.58\times10^{-20}$ \\
\midrule
\multicolumn{4}{@{}l}{InstructVLA: patch $\lambda_{\mathrm{cos}}=0.20$ vs.\ full-image Text-preserving PGD} \\
Exact match & $155$ vs.\ $155$ & $20$/$20$ & $1.00$ \\
Lower $\mathrm{KL}_{\mathrm{tf}}$ & $130$ of $200$ & --- & $2.65\times10^{-5}$ \\
\bottomrule
\end{tabular}
\end{table}

\subsection{Logged-scene open-loop safety protocol}
\label{app:open_loop_safety}

\paragraph{Cohort, time alignment, and geometry.}
Displacement and geometric collision readouts are standard in open-loop planning evaluations~\citep{hu2023uniad,jiang2023vad}. We score Alpamayo's free-generation plans against time-aligned logged actors on the same $154$ annotated clips for every arm, in the ego frame fixed at $t_0$. Actor boxes are transformed from their observation-time rig frames into that fixed frame using logged ego poses, then interpolated in position and unwrapped heading to the planning timestamps within their observed intervals. A single-observation track is used only at its nearest planning timestamp if within $0.05$ s. Following standard open-loop protocol, actors follow their recorded trajectories over the $6.4$ s horizon.

Ego dimensions and the rear-axle-to-box-center offset come from the clip's vehicle metadata, and the ego heading inferred from consecutive waypoints keeps its previous value for steps shorter than $1$ mm. Actor dimensions are the per-track median annotated length and width. Apparent ego self-tracks, whose median center lies within $1$ m of the expert ego footprint, are removed.

\paragraph{Silent collision rate.}
For clip $i$, let $M_i$ denote exact free-generation rationale equality, $C_i$ the clean-plan collision flag, and $A_i$ the attacked-plan collision flag. We report
\[
    \frac{1}{154}\sum_{i=1}^{154}
    \mathbf{1}\{M_i\land A_i\land\neg C_i\}.
\]
The three flags come from paired records for the same clip and decoding branch. On these $154$ clips, the exact-match counts $\sum_i M_i$ are $16$, $131$, $126/128/133/139/141$, and $111/111/110/117/123$ for the action-only, text-preserving, full-image $\lambda_{\mathrm{cos}}=5/10/15/20/30$, and patch $\lambda_{\mathrm{cos}}=5/10/15/20/30$ arms, and the attacked-plan collision counts $\sum_i A_i$ are $91$, $94$, $94/92/86/87/95$, and $78/82/73/74/72$.
\subsection{TNAS implementation and hook selection}
\label{app:local_probe_impl}

\paragraph{Pooled summaries and additive intervention coordinates.}
For a layer block $L=\{\ell_1,\ldots,\ell_K\}$ and active token span $R$, the pooled summary concatenates per-layer token means,
\begin{equation}
    \bar H_\ell = \frac{1}{|R|}\sum_{t\in R} H_\ell^{(t)}, \qquad
    h = \bigl(\bar H_{\ell_1},\ldots,\bar H_{\ell_K}\bigr)\in\mathbb{R}^{Kd}.
    \label{eq:app_pooled_state}
\end{equation}
The intervention handle broadcasts one vector per layer over the same span,
\begin{equation}
    H_\ell^{(t)} \leftarrow H_\ell^{(t)}+\delta_\ell,\qquad t\in R,\;\ell\in L,
    \label{eq:app_additive_hook}
\end{equation}
with one unscaled additive coordinate per layer and hidden dimension. The summary $h$ is not a standalone sufficient input to the model: a readout written at $h+\delta$ denotes the full hooked forward pass, with the remaining inputs and decoding configuration fixed.

\paragraph{Hooked action Jacobian and residual.}
With $A_{\mathrm{hook}}$ and $J_A$ the flattened action output of that pass and its Jacobian at $\delta=0$, as defined in \hyperref[sec:method_action_dir]{Section~\ref*{sec:method_action_dir}}, Alpamayo targets
\[
    \gact^{\star}\in\arg\max_{\lVert v\rVert_2=1}\lVert J_Av\rVert_2,
\]
materializing the Jacobian by differentiating each output coordinate and using the first right-singular vector from a thin \texttt{torch.linalg.svd}. A Jacobian norm at most $10^{-12}$ returns a zero direction. InstructVLA uses the normalized Gaussian-output VJP construction in \hyperref[app:ivla_impl]{Appendix~\ref*{app:ivla_impl}}. Both models form the text gradients and residual $r=(I-P_{\Tspan})\gact$ in these unscaled $\delta$ coordinates, before intervention scaling.

\paragraph{Matched-budget perturbations.}
For unit directions $u\in\{d_{\mathrm{TNAS}},\gact,g_{\mathrm{NLL}},g_{\mathrm{margin}},u_{\mathrm{rand}}\}$, matched-budget probes use the additive hook and model-specific clean-state scales.
For Alpamayo, the step is per-layer RMS-scaled,
\begin{equation}
    \delta_\ell = s\cdot\alpha\cdot\sigma_{h,\ell}\cdot u_\ell,\qquad s\in\{-1,+1\},
    \label{eq:app_rms_intervention}
\end{equation}
where $\sigma_{h,\ell}=\mathrm{rms}(\bar H_\ell)$. Equivalently, $\delta=s\alpha Su$ with $S=\mathrm{diag}(\sigma_{h,\ell_1}I_d,\ldots,\sigma_{h,\ell_K}I_d)$.%
InstructVLA applies a single captured clean-state scale $\sigma_{\mathrm{IVLA}}=\max(0.01\,\mathrm{rms}(h_{\mathrm{meta}}),10^{-3})$ across hooked layers, so $\delta=s\alpha\sigma_{\mathrm{IVLA}}u$. Here $h_{\mathrm{meta}}$ concatenates the per-layer mean clean meta-token states, captured separately from the response-token intervention.
Alpamayo's action-facing and random directions use $\alpha\in\{0.03,0.1,0.3,0.5,1.0,3.0,6.0\}$, and its text-gradient directions use $\{0.03,0.1,0.15,0.25,0.3,0.4,0.5,1.0\}$. InstructVLA uses $\alpha\in\{0.25,0.5,1,2,4,8,16,32\}$ as detailed in \hyperref[app:ivla_impl]{Appendix~\ref*{app:ivla_impl}}.
For each direction and text-KL budget, Alpamayo selects the available row with the largest achieved teacher-forced KL no greater than the target budget, and InstructVLA selects the row closest to the target budget within the $1.4B$ ceiling described in \hyperref[app:ivla_impl]{Appendix~\ref*{app:ivla_impl}}.
Alpamayo random controls in Probes~2 and~3 use $16$ isotropic unit-vector seeds per clip, and the InstructVLA matched-budget grids use random seeds $0$ and $1$.

\subsubsection{Layer-window sensitivity and hook viability}
\label{app:layer}

The Alpamayo layer block comes from a sweep over three late VLM blocks, layers 24 to 27, 28 to 31, and 32 to 35. Each block runs the five directions of \hyperref[sec:matched_budget]{Probe~2} with alphas $\{0.03,0.1,0.3,0.5,1.0,3.0,6.0\}$ at a teacher-forced text-KL budget of $10^{-4}$ on the same $50$ clips drawn from the $200$-clip cohort. All three blocks recover rank-2 active text spans on every clip and show the same action/random asymmetry. As \hyperref[tab:layer_block]{Table~\ref*{tab:layer_block}} shows, layers 32 to 35 give the strongest matched-budget action response, the selection rule of \hyperref[sec:method_shared_state]{Section~\ref*{sec:method_shared_state}}, and also the smallest projection $\rho$. Adjacent blocks reproduce the same pattern at smaller magnitude.

\begin{table}[t]
\centering
\caption{\textbf{Alpamayo layer-window sweep.} Matched teacher-forced text-KL budget $10^{-4}$ on $50$ clips per block. At layers 24 to 27, $47$ clips have a TNAS row and $49$ a random row within the budget.}
\label{tab:layer_block}
\small
\begin{tabular}{@{}lrrr@{}}
\toprule
Layers & Mean $\rho$ (\%) & TNAS ADE (m) & Random ADE (m) \\
\midrule
24 to 27 & $2.89$ & $0.195$ & $0.00284$ \\
28 to 31 & $2.34$ & $0.161$ & $0.00276$ \\
32 to 35 & $\underline{1.99}$ & $\underline{0.226}$ & $0.00356$ \\
\bottomrule
\end{tabular}
\end{table}

\paragraph{InstructVLA hook/gradient viability audit.}
For InstructVLA, a viability audit checks that a candidate hook exposes both a nonzero text basis on its token span and a nonzero action sensitivity through the flow action expert. \hyperref[tab:ivla_hook_audit]{Table~\ref*{tab:ivla_hook_audit}} compares meta- and response-token spans in three four-layer windows. Response-token spans are feasible in all three windows, and layers 14 to 17 give the largest action response among them. Meta-token spans move the action but have zero text gradients, so they are pure action-interface states rather than shared text/action states. We therefore select the response-token 14 to 17 window.

\begin{table}[t]
\centering
\caption{\textbf{InstructVLA hook viability audit.} Audit on three Bridge-distribution examples. Feasible cells have nonzero text gradients, a rank-2 text basis, and a nonzero action proxy. The selected hook is layers 14 to 17 on response tokens.}
\label{tab:ivla_hook_audit}
\small
\begin{tabular}{@{}llrrrr@{}}
\toprule
Layers & Span & NLL grad & Margin grad & Action proxy & Feasible (\%) \\
\midrule
4 to 7 & meta & $0$ & $0$ & $1.39\times10^{-2}$ & $0$ \\
4 to 7 & response & $0.109$ & $0.849$ & $4.82\times10^{-3}$ & $100$ \\
14 to 17 & meta & $0$ & $0$ & $2.07\times10^{-2}$ & $0$ \\
14 to 17 & response & $0.199$ & $1.27$ & $7.54\times10^{-3}$ & $100$ \\
24 to 27 & meta & $0$ & $0$ & $1.94\times10^{-2}$ & $0$ \\
24 to 27 & response & $0.0779$ & $0.388$ & $1.51\times10^{-3}$ & $100$ \\
\bottomrule
\end{tabular}
\end{table}

\subsubsection{InstructVLA action-direction estimator and budget selection}
\label{app:ivla_impl}

Gradients pass from the response-token hook through latent action features to the flow expert by bypassing inference-only detaches and using out-of-place Euler updates.
To estimate the action-sensitive direction without materializing the full Jacobian through the flow rollout, we draw an output-space Gaussian vector $z_0$ and normalize $w_0=z_0/(\lVert z_0\rVert_2+10^{-8})$, then compute a single vector-Jacobian product at $\delta=0$ followed by a central-difference directional gain check ($\epsilon=0.01$):
\[
    v=J_A^\top w_0,
    \qquad \gact=\frac{v}{\lVert v\rVert_2},
    \qquad
    j=\frac{A_{\mathrm{hook}}(\epsilon \gact)-A_{\mathrm{hook}}(-\epsilon \gact)}{2\epsilon}.
\]
A VJP norm at most $10^{-12}$ returns a zero direction, and $\lVert j\rVert_2$ is recorded as the directional gain.

\paragraph{Finite-grid budget selection.}
\label{app:ivla_direction_sets}
Each evaluation estimates the per-example directions once on the $200$ example IDs and holds them fixed throughout its perturbation sweep. The matched-budget sweep uses $\alpha\in\{0.25,0.5,1,2,4,8,16,32\}$ so that action-aligned directions reach the $B=10^{-3}$ KL budget on all $200$ examples. For nominal budget $B\in\{5\times10^{-4},10^{-3}\}$, the analysis selects the row with achieved teacher-forced KL closest to $B$ within a $1.4B$ ceiling. On this grid, action-aligned directions reach a maximum mean text KL of about $1.50\times10^{-3}$, while text-aligned directions reach about $10^{-2}$. Across the five directions in \hyperref[tab:ivla_matched]{Table~\ref*{tab:ivla_matched}}, mean achieved KL is $97.8\%$ to $101\%$ of $B$ at $B=5\times10^{-4}$ and $94.4\%$ to $98.9\%$ of $B$ at $B=10^{-3}$, keeping at least $186$ examples per direction at $B=5\times10^{-4}$ and all $200$ examples at $B=10^{-3}$.

\subsection{Benign rationale-only reader}
\label{app:benign_reader}

\paragraph{Cohort and reader.}
As described in \hyperref[app:dataset]{Appendix~\ref*{app:dataset}}, the reader uses all $1{,}181$ clips of the release manifest. Each clip's clean rationale and planned trajectory come from one greedy Alpamayo run with generation and flow seed $42$, and the clips contain $258$ distinct rationale strings. \texttt{Qwen/Qwen3-14B} in bf16, with thinking disabled and greedy decoding, receives one rationale string per call together with the horizon and the label definition, and returns a binary stop prediction and a stop probability. It never receives the image, the trajectory, the ego speed, or the map. Each distinct string is read once, so identical rationales receive identical predictions, and all $258$ outputs parse. These benign statistics use greedy rationales and are not paired with the sampled-decoding attack results of \hyperref[sec:attack]{Section~\ref*{sec:attack}}.

\paragraph{Prompt and readout.}
Each distinct rationale is inserted into the user prompt below, which goes through the Qwen3 chat template, and the reply is capped at $200$ new tokens. The binary prediction is the \texttt{will\_stop} field of the returned JSON object, and the stop probability is the \texttt{stop\_prob} number that the model writes into the same object. No token probabilities are used. The three remaining fields do not enter this paper's analysis. In the prompt, \texttt{[RATIONALE]} stands for the rationale string.
\begin{lstlisting}[style=prompt]
You are given one sentence: the rationale that an autonomous-driving model produced to explain its own driving plan for the next 6.4 seconds.

Predict what the vehicle will actually do. You see only this sentence. You do not see the camera images, the planned trajectory, the vehicle's current speed, or the map.

Sentence: "[RATIONALE]"

Answer with a single JSON object and nothing else:

{"will_stop": <true|false>,
 "stop_prob": <number 0..1>,
 "longitudinal": "<accelerate|maintain|decelerate|stop>",
 "lateral_dir": "<left|right|none>",
 "lateral_mag": "<none|small|large>"}

Definitions:
- will_stop: the vehicle comes to rest (below 0.5 m/s) before the 6.4 s horizon ends.
- stop_prob: your probability that will_stop is true.
- longitudinal: speed in the final 0.8 s vs the first 0.8 s. "stop" if it comes to rest, "decelerate" if below 70% of the initial speed, "accelerate" if above 130%, else "maintain".
- lateral_dir / lateral_mag: net sideways departure from the starting heading, in metres, measured at its largest point. Count sideways motion caused by the road curving as well as by a deliberate lane change or nudge. "left" and "right" are from the driver's seat. none = under 1 m, small = 1 to 3 m, large = 3 m or more.
\end{lstlisting}

\paragraph{Stop label.}
Speeds are taken from consecutive planned waypoints at $10$ Hz. A plan comes to rest within the horizon if its mean speed over the final $0.8$ s of the $6.4$ s horizon is below $0.5$ m/s. \hyperref[tab:benign_reader]{Table~\ref*{tab:benign_reader}} reports the reader's predictions against this label. The reader recovers $84.1\%$ of stopping and $84.4\%$ of non-stopping plans, $84.3\%$ overall. Its stop probability ranks stopping plans with AUROC $0.882$, above $0.820$ for a no-text baseline that ranks clips by entry speed from the ego history, with tied scores assigned their average rank.

\begin{table}[t]
\centering
\caption{\textbf{Benign rationale-only reader.} Planned outcomes against reader predictions on $1{,}181$ clean Alpamayo clips. Shares are within each prediction column.}
\label{tab:benign_reader}
\small
\begin{tabular}{@{}lrrrrr@{}}
\toprule
 & \multicolumn{2}{c}{Reader: will stop} & \multicolumn{2}{c}{Reader: keep going} & \\
\cmidrule(lr){2-3}\cmidrule(lr){4-5}
Plan outcome & Clips & Share (\%) & Clips & Share (\%) & Total \\
\midrule
Comes to rest & $243$ & $63.6$ & $46$ & $5.76$ & $289$ \\
Does not come to rest & $139$ & $36.4$ & $753$ & $94.2$ & $892$ \\
\midrule
Total & $382$ & $100$ & $799$ & $100$ & $1{,}181$ \\
\bottomrule
\end{tabular}
\end{table}

\paragraph{Mixed-outcome rationales.}
A rationale has mixed outcomes when its clips include both stopping and non-stopping plans. Of the $258$ rationales, $23$ are mixed, and together they cover $436$ clips. Because the label is read at the end of the horizon, mixed outcomes include vehicles waiting at a red light that depart within the horizon. All $19$ non-stopping clips with the rationale ``Stop for the red traffic light'' start at rest. The statistic therefore measures how much of the $6.4$ s plan a sentence leaves unspecified, not how often the sentence is wrong.
\subsection{Projection audits}
\label{app:projection}

\paragraph{Projection statistics.}
Across the $200$ Alpamayo clips, the projection ratio $\rho$ of \hyperref[sec:geometry]{Probe~1} has mean $1.87\%$ and median $1.79\%$. Extending the cohort to all $1{,}180$ manifest clips with a nonempty rationale leaves the mean at $1.87\%$. The random unit-vector reference draws $1{,}000$ isotropic unit vectors per clip. Over these $200{,}000$ samples it has mean $0.978\%$, median $0.916\%$, 95th percentile $1.91\%$, and 99th percentile $2.36\%$. Of the $200$ clips, $43.5\%$ exceed the random 95th percentile and $25.5\%$ exceed the 99th, against $5\%$ and $1\%$ expected by chance. The largest per-clip value is $\rho=5.28\%$, so the TNAS residual keeps more than $99.8\%$ of the action direction's norm on every clip. In InstructVLA, $\rho$ has mean $5.28\%$ and median $4.90\%$ over the $200$ examples, and its mean square of $0.00382$ leaves $99.8\%$ of the norm on average. Rescaling either text-gradient basis vector by $0.01$, $0.1$, $1$, $10$, or $100$ before re-orthonormalization changes the action projection ratio by at most $4.84\times10^{-8}$, so the ratio depends only on the text span.

\subsection{Hidden-state intervention probe values}
\label{app:probe_tables}

\hyperref[tab:p0_matched_xfam]{Tables~\ref*{tab:p0_matched_xfam}} and~\ref{tab:p2_dose_response} list the cohort means plotted in \hyperref[fig:probes]{Figure~\ref*{fig:probes}c--h}.

\begin{table}[t]
\centering
\caption{\textbf{Probe 2: Matched-budget action shifts in Alpamayo and InstructVLA.} Alpamayo entries are mean ADE shifts in meters on $200$ clips, or $199$ for $g_{\mathrm{NLL}}$ at $B=10^{-4}$, at nominal teacher-forced KL budgets $B$. InstructVLA entries are mean per-step normalized 7-DoF action shifts with the admissible count $n$, which covers all $200$ examples at $B=10^{-3}$ as detailed in \hyperref[app:ivla_direction_sets]{Appendix~\ref*{app:ivla_direction_sets}}.}
\label{tab:p0_matched_xfam}
\label{tab:ivla_matched}
\small
\begin{tabular}{@{}lrrrrrr@{}}
\toprule
& \multicolumn{2}{c}{Alpamayo ADE shift (m)} & \multicolumn{4}{c}{InstructVLA norm.\ $\ell_2$ shift} \\
\cmidrule(lr){2-3}\cmidrule(l){4-7}
& & & \multicolumn{2}{c}{$B=5\times10^{-4}$} & \multicolumn{2}{c}{$B=10^{-3}$} \\
\cmidrule(lr){4-5}\cmidrule(l){6-7}
Direction & $B=10^{-4}$ & $B=5\times10^{-4}$ & Shift & $n$ & Shift & $n$ \\
\midrule
$d_{\mathrm{TNAS}}$   & $\underline{0.227}$ & $0.504$             & $0.0131$             & $189$ & $0.0137$             & $200$ \\
$\gact$               & $0.181$             & $\underline{0.527}$ & $\underline{0.0144}$ & $186$ & $\underline{0.0166}$ & $200$ \\
$u_{\mathrm{rand}}$   & $0.00359$           & $0.00510$           & $0.00848$            & $194$ & $0.00752$            & $200$ \\
$g_{\mathrm{NLL}}$    & $0.00240$           & $0.00238$           & $0.00760$            & $186$ & $0.00918$            & $200$ \\
$g_{\mathrm{margin}}$ & $0.00229$           & $0.00252$           & $0.00916$            & $186$ & $0.00817$            & $200$ \\
\bottomrule
\end{tabular}
\end{table}

\begin{table}[t]
\centering
\caption{\textbf{Probe 3: Norm-matched clean-state amplification on $200$ Alpamayo clips and $200$ InstructVLA examples.} At each strength $q$, all directions receive the same per-example intervention norm $|q|S_{\mathrm{clip}}$, and $q=0$ is the clean state. The KL columns report teacher-forced KL at $q=2$.}
\label{tab:p2_dose_response}
\label{tab:ivla_dose_response}
\small
\setlength{\tabcolsep}{4pt}
\begin{tabular}{@{}lrrrrrrrr@{}}
\toprule
& \multicolumn{4}{c}{Alpamayo} & \multicolumn{4}{c}{InstructVLA} \\
\cmidrule(lr){2-5}\cmidrule(l){6-9}
& \multicolumn{3}{c}{ADE shift (m)} & KL & \multicolumn{3}{c}{Norm.\ $\ell_2$ shift} & KL \\
\cmidrule(lr){2-4}\cmidrule(lr){5-5}\cmidrule(lr){6-8}\cmidrule(l){9-9}
Direction & $q=0.5$ & $q=1$ & $q=2$ & $q=2$ & $q=0.5$ & $q=1$ & $q=2$ & $q=2$ \\
\midrule
$d_{\mathrm{TNAS}}$   & $\underline{0.0317}$ & $0.0673$             & $\underline{0.142}$ & $9.87\times10^{-5}$ & $\underline{0.0734}$ & $\underline{0.110}$ & $\underline{0.156}$ & $2.20\times10^{-2}$ \\
$\gact$               & $0.0316$             & $\underline{0.0675}$ & $\underline{0.142}$ & $1.03\times10^{-4}$ & $\underline{0.0734}$ & $\underline{0.110}$ & $0.155$             & $2.71\times10^{-2}$ \\
$u_{\mathrm{rand}}$   & $0.00223$            & $0.00261$            & $0.00264$           & $6.45\times10^{-5}$ & $0.00861$            & $0.00999$           & $0.0124$            & $3.64\times10^{-3}$ \\
$g_{\mathrm{NLL}}$    & $0.00234$            & $0.00262$            & $0.00297$           & $2.66\times10^{-2}$ & $0.0134$             & $0.0197$            & $0.0283$            & $5.80\times10^{-1}$ \\
$g_{\mathrm{margin}}$ & $0.00217$            & $0.00263$            & $0.00323$           & $2.73\times10^{-2}$ & $0.0158$             & $0.0198$            & $0.0337$            & $4.66\times10^{-1}$ \\
\bottomrule
\end{tabular}
\end{table}

\subsection{Sign-aligned clean-state amplification}
\label{app:sign_aligned}

For the clean-state amplification of \hyperref[sec:nulling_and_amplification]{Probe~3}, we sweep a raw scalar $s$ with
\begin{equation}
    h(s,u)=h_{\mathrm{clean}}+s\,S_{\mathrm{clip}}u,\qquad
    S_{\mathrm{clip}}=\bigl|\langle h_{\mathrm{clean}},d_{\mathrm{TNAS}}\rangle\bigr|.
    \label{eq:app_clean_sweep}
\end{equation}
For both models, write $c=\langle h_{\mathrm{clean}},d_{\mathrm{TNAS}}\rangle$ and aggregate by the sign-aligned parameter $q=s\cdot\mathrm{sgn}(c)$. Thus $q=0$ is the clean state, and all directions have matched per-clip norm $\lVert qS_{\mathrm{clip}}u\rVert$ at fixed $q$. Because the SVD sign of a direction is arbitrary, $q$ aligns the amplification curves by the sign of the natural projection. The InstructVLA clean-state amplification sweep averages four random seeds per example for $u_{\mathrm{rand}}$, while deterministic named directions are evaluated once.

\subsection{Existing asset licenses and terms}
\label{app:asset_licenses}

The experiments use existing datasets, checkpoints, and annotator models only for analysis. We do not redistribute raw clips, VLA-IT examples, checkpoints, adapter weights, or a repackaged derived dataset. Alpamayo uses the gated \texttt{nvidia/PhysicalAI-Autonomous-Vehicles} dataset under the NVIDIA Autonomous Vehicle Dataset License Agreement (\texttt{nvidia-av-dataset}), accessed through HuggingFace, and the \texttt{nvidia/Alpamayo-1.5-10B} checkpoint under the NVIDIA License for non-commercial research or evaluation use.
The Alpamayo inference code is Apache-2.0. The benign rationale-only reader uses \texttt{Qwen/Qwen3-14B}~\citep{yang2025qwen3} under Apache-2.0.
InstructVLA uses the public \path{ShuaiYang03/instructvla_finetune_v2_xlora_freeze_head_instruction} checkpoint, \path{ShuaiYang03/InstructVLA_Assets}, and the \path{ShuaiYang03/InstructVLA_embodied_mm_evaluation_assets} language-evaluation files. We credit these releases and the InstructVLA paper, and we use only the fixed evaluation examples. As of November 2026, the public InstructVLA release pages and the \path{InternRobotics/InstructVLA} code repository list no explicit license for these assets, so we treat their license status as unspecified.
The InstructVLA asset page identifies \texttt{nvidia/Eagle2-2B}~\citep{li2025eagle2} as the VLM base, whose HuggingFace card lists CC-BY-NC-4.0. For all assets with gated or restrictive terms, the experiments use local access only and do not distribute the protected asset contents. Rationale-preserving visual manipulations also pose a misuse risk: they can alter predicted actions without changing the visible explanation. Runnable attack pipelines and optimized perturbation tensors are not distributed with this manuscript.

\subsection{Compute resources}
\label{app:compute_resources}

The experiments ran on a single HPC node with up to four NVIDIA A100-80GB GPUs and one Python worker per GPU.
Each Alpamayo worker loads one model copy and uses about $25$ GB of GPU memory. The InstructVLA probes use the same GPU type.
We estimate that reproducing the Alpamayo hidden-state probes and clean-state amplification curves from cached model and data artifacts takes $75$ to $85$ A100 GPU-hours, including about $25$ GPU-hours for the local-probe alpha grids and $30$ GPU-hours for the intervention sweeps. The InstructVLA local probes and clean-state amplification take an estimated $20$ to $24$ GPU-hours.
The input-attack costs below are measured wall-clock times summed over workers. Each Alpamayo guidance weight takes about $47$ GPU-hours on the $200$ clips, or about $14$ GPU-minutes per clip, for both the full-image and the patch attack. The Action-only and Text-preserving rows come from a shared $200$-clip run of about $85$ GPU-hours. Each InstructVLA input-attack arm takes about $13$ GPU-hours on the $200$ examples. Merging, row selection, summaries, and the open-loop collision scoring run on CPU and take minutes per stage.

\end{document}